\documentclass{article}
\usepackage[preprint]{corl_2026}
\usepackage{graphicx}
\usepackage{subcaption}
\usepackage{amsmath}
\usepackage{amsfonts}
\usepackage{wrapfig}
\usepackage{multirow}
\usepackage{booktabs}
\usepackage{makecell}
\usepackage{float}
\usepackage{hyperref}

\title{ORD-WM: Learning Object-Centric Rigid Dynamics for 3D World Modeling in Robotic Manipulation}

\author{
Yueyang Weng$^{*}$,
Xiaopeng Zhang$^{*}$,
Yuxi Ou,
Yongjin Mu,
Chengzhi Zhao,
Yanjie Li$^{\ddagger}$
\\[6pt]
Harbin Institute of Technology, Shenzhen
\\[6pt]
$^{*}$Equal contribution, \quad $^{\ddagger}$Corresponding author
\\[4pt]
\url{https://hitsz-wyy.github.io/ORD-WM/}
}

\hypersetup{
  pdftitle={ORD-WM: Learning Object-Centric Rigid Dynamics for 3D World Modeling in Robotic Manipulation},
  pdfauthor={Yueyang Weng, Xiaopeng Zhang, Yuxi Ou, Yongjin Mu, Chengzhi Zhao, Yanjie Li},
  pdfsubject={Preprint},
  pdfkeywords={World Models, Robot Manipulation, 3D Dynamics Prediction}
}

\begin{document}
\maketitle

\begin{figure*}[h]
    \vspace{-10pt}
    \centering
    \includegraphics[width=0.98\textwidth]{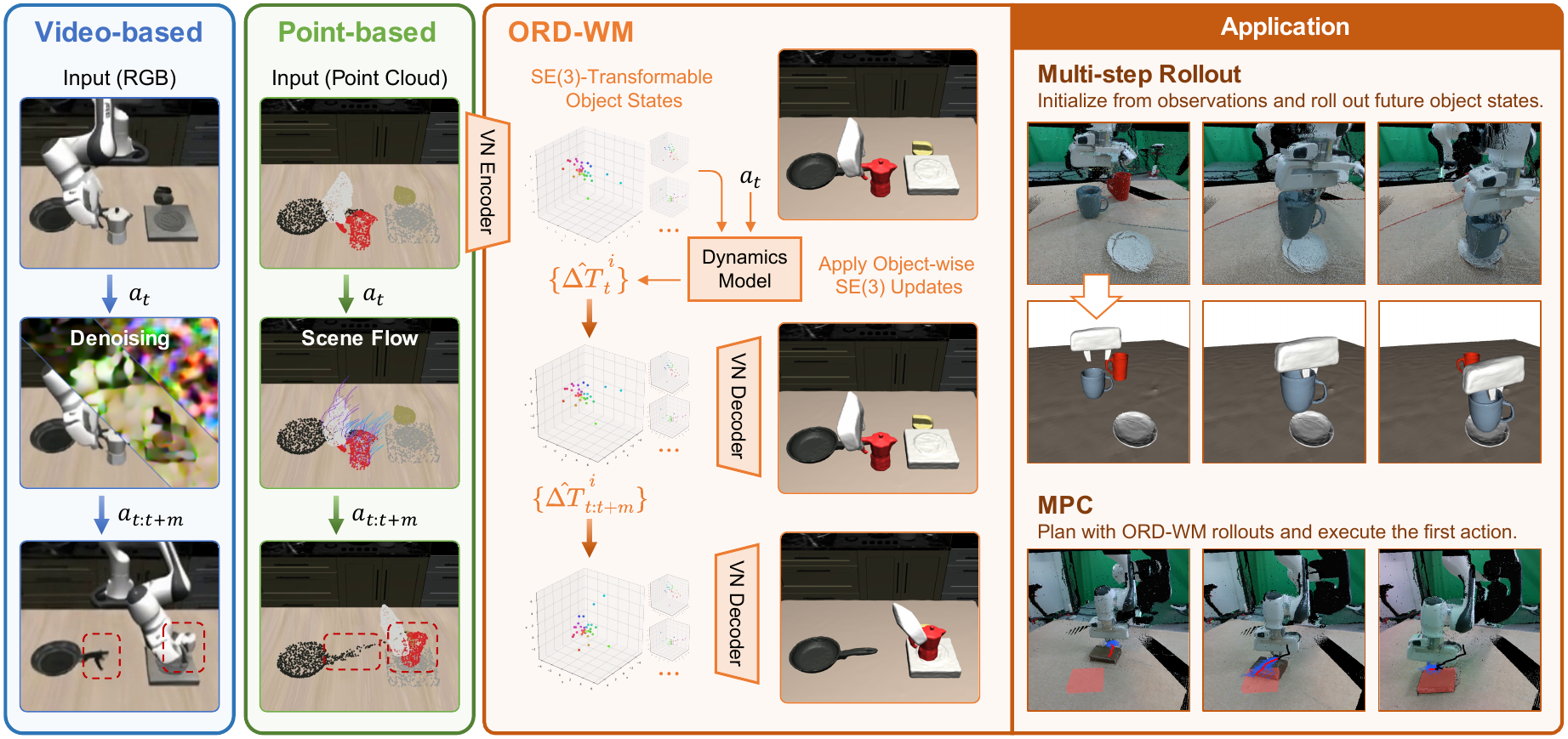}
    \caption{In contrast to video-based rollout and point-based scene-flow prediction, which may suffer from visual artifacts, shape drift, or object-level inconsistency, ORD-WM performs multi-step prediction by analytically applying predicted object-wise SE(3) updates to geometry-aware object latents composed of 3D vector features, preserving object geometry while updating pose. This latent-space rollout enables real-world object motion prediction and closed-loop model predictive control, where candidate actions are evaluated by their predicted effects on object states.
    }
    \label{fig:overall}
\end{figure*}

\begin{abstract} 
World models provide robots with a predictive mechanism for reasoning about how actions change object states in manipulation tasks. However, existing image-based and dense 3D world models do not explicitly impose object-level rigid motion, leaving rigid-object geometry preservation to be learned implicitly and making multi-step rollouts susceptible to shape drift or object-level motion inconsistency. To address this, we propose ORD-WM, a 3D world model that learns object-centric rigid dynamics for robotic manipulation. ORD-WM learns a vector-neuron variational autoencoder supervised by signed distance fields, which encodes each object point cloud as a geometry-aware latent representation composed of 3D vector features, and uses an action-conditioned dynamics model to predict object-wise SE(3) motion increments. Rather than generating unconstrained next latent states, ORD-WM analytically applies predicted object-wise rigid transformations to the current latent states, separating persistent object geometry from pose changes. Experiments show that ORD-WM enables stable, geometry-preserving multi-step prediction and supports efficient real-time replanning for closed-loop model predictive control in simulated and real-world manipulation scenarios.

\end{abstract}


\section{Introduction}

    Humans can anticipate how physical scenes evolve and how actions change object states, which is fundamental to everyday interaction with the physical world. Endowing robots with a similar predictive ability is an important step toward more capable manipulation: before pushing, grasping, or rearranging objects, a robot should reason about how its actions may move and transform the objects in a scene. World models provide a natural framework for acquiring such capability by learning action-conditioned dynamics and rolling out future states internally, enabling robots to evaluate candidate actions and plan toward desired object configurations without relying solely on trial-and-error interaction in the real world~\citep{ha2018world,hafner2025mastering}.
    
    Recent robotic world models have explored several forms of action-conditioned prediction. Video-based models roll out future visual observations and can capture rich visual changes in robot-object interaction~\citep{mei2026video}. 
    However, because prediction is performed in image or visual latent space, these models lack an explicit metric 3D state. As a result, visual plausibility alone does not guarantee temporally consistent object geometry or physically plausible interactions during rollout~\citep{jiang2026robowm,zhou2026toward}.
    Dense 3D world models based on point clouds, particles, or Gaussian scene representations move prediction closer to physical space by forecasting the motion of 3D elements~\citep{ai2025review}. However, when modeling rigid object motion through local element-wise updates, these models must coordinate local motions to produce globally consistent object transformations, which may accumulate into shape distortions or object-level drift during multi-step rollout~\citep{huang2026pointworld}.

	For manipulation tasks involving rigid or approximately rigid objects, prediction should preserve object geometry while allowing object poses to change through interaction. Rigid-body simulators~\citep{todorov2012mujoco} offer a useful abstraction for this setting: they represent scenes as collections of bodies with persistent geometry and update object states through constrained body-level motion, rather than by independently moving dense visual or geometric elements. This separation between geometry and motion suggests a natural inductive bias for learned world models of rigid-object manipulation.

    Building on this perspective, we propose ORD-WM, a 3D world model that learns object-centric rigid dynamics for robotic manipulation (Fig.~\ref{fig:overall}). 
    ORD-WM represents a manipulation scene as a collection of geometry-aware object states and evolves them through object-wise rigid motion. To obtain such states, ORD-WM trains a signed-distance-field (SDF)-supervised vector-neuron (VN)~\citep{deng2021vector} variational autoencoder (VAE)~\citep{kingma2013auto}, denoted SDF-VNVAE. Its encoder maps each object point cloud to a geometry-aware latent state composed of 3D vector features that capture object geometry and can be analytically transformed by rigid motions in latent space. Instead of predicting unconstrained next latent states, ORD-WM predicts object-wise SE(3)\footnote{SE(3) denotes the Special Euclidean group in three dimensions, representing 3D rigid transformations consisting of a rotation $R \in SO(3)$ and a translation $t \in \mathbb{R}^3$.} motion increments conditioned on the current scene state and robot action, and applies these transformations directly to the object states.
	
    Together, these design choices yield an object-centric 3D world model in which object geometry is maintained as a persistent latent structure and future states are generated through structured rigid updates. We evaluate ORD-WM on multi-step dynamics prediction and model predictive control (MPC) for object pushing and pick-and-place, showing that its object-centric SE(3)-structured rollout improves geometric consistency and supports downstream model-based control.

\section{Related Work}
\label{sec:related}

	\textbf{World Models for Robotic Manipulation.} 
	World models~\citep{ha2018world} learn action-conditioned predictions of future states or observations, providing a predictive substrate for planning and policy learning in imagination~\citep{hafner2025mastering,hafner2019learning,hafnerdream,hafner2025training}. 
    Recent robotic world models extend this paradigm by learning action-conditioned predictors from large-scale video and robot interaction data. These models forecast future images~\citep{wuunleashing,zhou2024robodreamer,Zhu_2025_ICCV} or visual latents~\citep{bardesrevisiting,assran2025v}, supporting policy learning~\citep{wu2023daydreamer,zhu2025wmpo}, policy conditioning~\citep{hu2025video}, visual model-based control~\citep{finn2017deep,ebert2018visual}, and policy evaluation~\citep{quevedo2025evaluating,tseng2025scalable}.
    However, prediction in image or visual-latent spaces does not by itself impose physical consistency: visually plausible rollouts may still fail to preserve object geometry, identity, or rigid-body motion over time. Another line of work grounds prediction in 3D, using point clouds~\citep{huang2026pointworld,huangparticleformer}, particles~\citep{li2019learning,sanchez2018graph}, or Gaussian scene representations~\citep{lu2025gwm,lu2024manigaussian} to model physical scene evolution. While these representations move prediction closer to physical space~\citep{xu2020learning}, dense element-wise motion still leaves rigid consistency to be learned implicitly: all points of a rigid object must agree with a single object-level transformation~\citep{byravan2017se3}. ORD-WM targets this rigid-object regime by evolving object states through object-wise SE(3) motion increments rather than future frames or dense point flows.

	\textbf{Object-Centric Representations for Dynamics Modeling.}
	Object-centric representations factorize a scene into entities rather than representing the entire scene with a single holistic latent, providing a natural structure for learning compositional dynamics~\citep{ai2025review,battaglia2016interaction,watters2017visual,kipf2020contrastive}. This factorization mirrors physical scenes, where objects move independently, interact relationally, and form new configurations across contexts~\citep{veerapaneni2020entity}. Object-centric dynamics have been shown to benefit prediction~\citep{watters2017visual}, planning~\citep{ye2020object}, model-based reinforcement learning~\citep{ferraro2025focus,wang2025dyno}, and relation-aware manipulation~\citep{mosbach2025sold} by making dynamics more modular and interpretable. However, existing methods typically use abstract object representations that support compositional reasoning but do not explicitly encode object geometry or constrain transitions to follow rigid-body motion~\citep{kipf2020contrastive,mosbach2025sold,locatello2020object}. As a result, rigid geometry preservation remains an implicit property to be learned from data. ORD-WM builds on this object-centric view, but represents each object with a geometry-aware latent state composed of 3D vector features and updates the latent state through explicit rigid transformations, thereby imposing rigid structure directly on the latent transition.

	\textbf{Implicit 3D Object Representations.}
    Implicit 3D object representations, such as SDFs and occupancy fields, model object shape as continuous functions and provide geometric information beyond discrete surface samples~\citep{jones20063d,xu2020signed}. 
    In robotics, SDFs have been widely used for collision-aware motion planning, where reasoning about geometry and spatial proximity is essential~\citep{zucker2013chomp,finean2021predicted,vasilopoulos2023ramp,driess2022learning}. 
    Learning-based SDF methods use signed distance supervision to learn object shape representations for reconstruction~\citep{park2019deepsdf}, perception~\citep{van2020learning,song2025implicit,yan2023implicit}, and manipulation planning~\citep{khargonkar2023neuralgrasps,ha2022deep}.
    ORD-WM trains a VNVAE with an SDF reconstruction objective, producing geometry-aware latents composed of 3D vector features that serve as object states for rigid, geometry-preserving rollout.

\section{Method}

We present ORD-WM, a 3D world model that learns object-centric rigid dynamics for robotic manipulation. ORD-WM represents a manipulation scene as a collection of geometry-aware object states and evolves them through object-wise rigid motion. Given RGB-D observations, ORD-WM first constructs object-level point clouds and encodes each object into a geometry-aware latent state composed of 3D vector features. An action-conditioned dynamics model then predicts object-wise SE(3) motion increments, which are analytically applied to the current latent states to enable structured multi-step rollout while preserving object geometry and identity.

\subsection{Object-Centric 3D Observation}
\label{sec:method_obs}

ORD-WM operates on object-centric 3D observations constructed from multi-view RGB-D inputs.
At each time step, the robot receives observations from $V$ camera views,
\begin{equation}
o_t=\{(I_t^v,D_t^v)\}_{v=1}^{V},
\end{equation}
where $I_t^v$ and $D_t^v$ denote the RGB image and aligned depth map from view $v$. 

We use Segment Anything Model~3 (SAM~3)~\citep{carion2025sam3segmentconcepts} to extract object masks in each RGB view, apply the masks to the corresponding depth maps, and back-project the masked depth pixels into a common world frame. 
Partial point clouds of the same object instance are merged across views, yielding
\begin{equation}
\mathcal{P}_t=\{P_t^i\}_{i=1}^{N_t},
\end{equation}
where $P_t^i$ is the point cloud of object $i$ and $N_t$ is the number of detected object instances.

\subsection{Geometry-Aware Object Representation Learning}

\begin{figure}[t]
    \centering
    \includegraphics[width=0.95\textwidth]{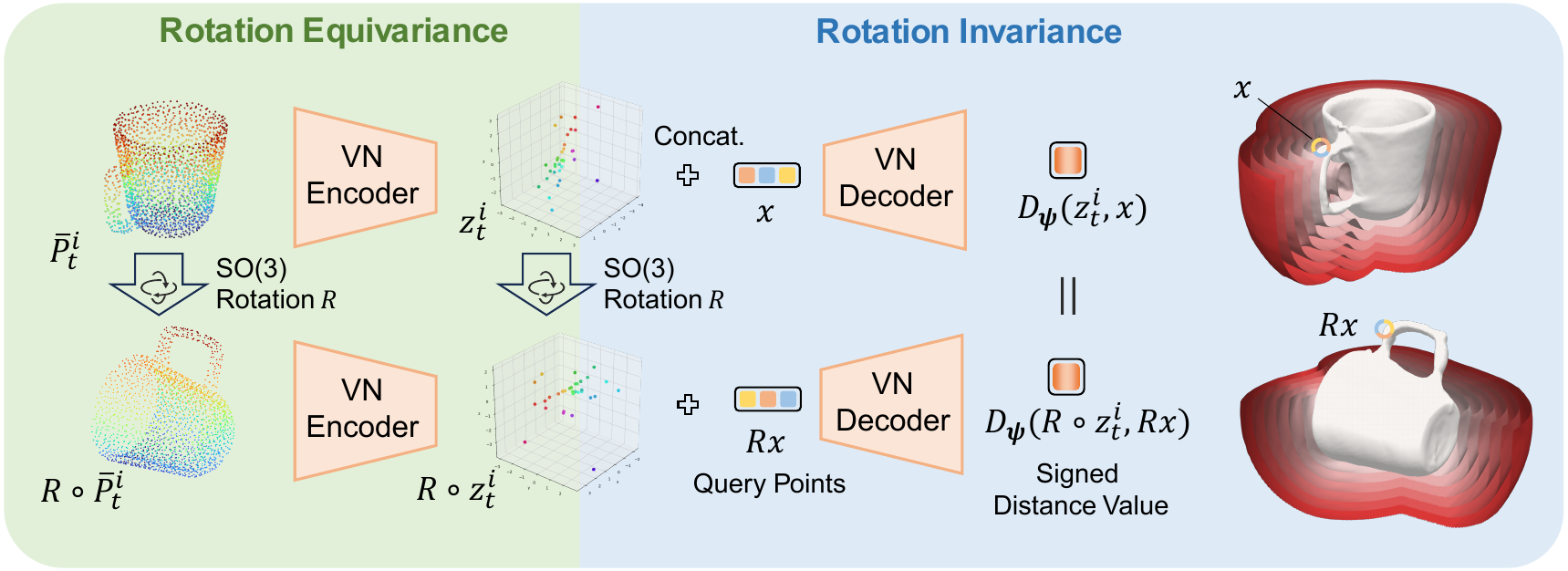}
    \caption{Overview of SDF-VNVAE. The VN encoder maps an object-centered point cloud to a latent representation composed of 3D vector features that rotate equivariantly with the input. The decoder predicts signed distance values from the latent sample $z_t^i$ and query points, and is invariant under their shared 3D rotation. For visual simplicity, the encoder output is drawn as $z_t^i$; during training, $z_t^i$ is sampled from the posterior parameterized by $(\mu_t^i,\sigma_t^i)$.
}
    \vspace{-6pt}
    \label{fig:vnvae}
\end{figure}

To support structured rigid dynamics in latent space, ORD-WM requires object states that encode geometry while allowing rotations to be applied analytically. We therefore train a geometry-aware VNVAE that maps each object point cloud to a latent representation composed of 3D vector features. The key property of this representation is SO(3)-equivariance: rotating the input object induces the same rotation on each 3D vector feature, making it meaningful to apply predicted object rotations directly to the latent representation. The decoder predicts signed distance values from the latent representation and query points in a rotation-invariant manner, encouraging the latent to capture object geometry and surrounding distance structure, as illustrated in Fig.~\ref{fig:vnvae}.

Before encoding, each object point cloud is centered by subtracting its centroid:
\begin{equation}
\bar{P}_t^i=\{p-c_t^i \mid p\in P_t^i\},
\end{equation}
where $c_t^i\in\mathbb{R}^3$ denotes the centroid of $P_t^i$.
For each centered point cloud $\bar{P}_t^i$, we use a VN-PointNet encoder $E_\phi$~\citep{deng2021vector} to parameterize a latent distribution over 3D vector features: 
\begin{equation}
(\mu_t^i,\sigma_t^i)=E_\phi(\bar{P}_t^i),
\qquad
\mu_t^i\in\mathbb{R}^{K\times 3}, \quad \sigma_t^i\in\mathbb{R}^{K}.
\end{equation}
The mean $\mu_t^i$ is a set of three-dimensional vectors, while $\sigma_t^i$ parameterizes one isotropic uncertainty value per vector feature. By construction, the mean vectors transform equivariantly under rotations:
\begin{equation}
\mu(R\circ \bar{P}_t^i)=R\circ \mu(\bar{P}_t^i), \qquad R\in SO(3),
\end{equation}
where $R\circ X$ denotes applying rotation $R$ to every point or 3D vector in $X$. This equivariance allows the dynamics model to apply predicted object rotations directly to the latent representation.

The decoder represents each object as an implicit SDF conditioned on the object latent. During representation learning, we sample $z_t^i$ from the posterior parameterized by $(\mu_t^i,\sigma_t^i)$ and query the decoder at a point $x\in\mathbb{R}^3$ in the object-centered frame to predict the signed distance value:
\begin{equation}
\hat{d}=D_\psi(z_t^i,x).
\end{equation}
We design the decoder to be invariant under a shared 3D rotation of the latent representation and query point. Together with encoder equivariance, this preserves the decoded geometry in the corresponding rotated frame.

We train the VNVAE with an SDF reconstruction objective, augmented by Eikonal~\citep{gropp2020implicit} and Kullback--Leibler (KL) regularization terms~\citep{kingma2013auto}. After training, we use the posterior mean $\mu_t^i$ as the deterministic geometry latent and append the retained object center $c_t^i$ as an additional 3D vector in the object state:
\begin{equation}
s_t^i=[\mu_t^i;c_t^i]\in\mathbb{R}^{(K+1)\times 3},
\qquad
S_t=\{s_t^i\}_{i=1}^{N_t}.
\label{eq:object-state}
\end{equation}
The full posterior parameterization, transformation properties, architecture, loss definitions and training details are provided in Appendix~\ref{app:vnvae}.

\subsection{Object-Centric Rigid Dynamics}

\begin{wrapfigure}{r}{0.44\textwidth}
    \vspace{-14pt}
    \centering
    \includegraphics[width=0.9\linewidth]{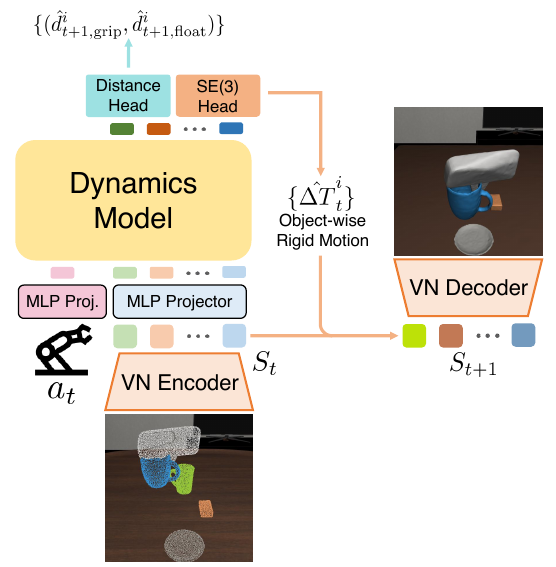}
    \caption{Overview of the dynamics model. The current scene state, robot proprioception, and action are encoded into tokens, which are processed to predict object-wise rigid motions and update the scene state.}
    \label{fig:dynamic_model}
\end{wrapfigure}

Given the object-centric scene state $S_t$, robot proprioception $\rho_t$, and robot action $a_t$, ORD-WM learns a dynamics model that predicts action-conditioned object motion in the scene (Fig.~\ref{fig:dynamic_model}). 
We partition scene objects into floating objects, whose poses can change over time, and fixed objects, which provide static context such as support surfaces.
Robot proprioception is converted into geometry-aware gripper tokens by applying forward kinematics to pre-sampled points on each gripper part, which are then encoded with the same VNVAE encoder used for scene objects.
Instead of using a single end-effector pose and gripper width, we represent the action as rigid transformations of individual gripper parts. 
Floating objects, fixed objects, gripper parts, and actions are tokenized with learned role embeddings and fed into a transformer dynamics module, which captures robot-object and object-object interactions and produces interaction-aware features $h_t^i$ for each scene object.

For each floating object feature, an SE(3) motion head predicts a translation increment and a 6D rotation representation~\citep{zhou2019continuity}: 
\begin{equation}
(\Delta \hat{x}^i_t, \hat{r}^i_t) = \mathrm{Head}_{\mathrm{se3}}(h^i_t) \quad \text{for } i \in \mathcal{F}_t,
\end{equation}
where $\mathcal{F}_t$ denotes the set of floating-object indices.
The rotation output is mapped to a valid rotation matrix $\hat{\Delta R}_t^i=\Pi_{\mathrm{SO(3)}}(\hat{r}_t^i)$, yielding the object-wise rigid motion $\hat{\Delta T}_t^i=(\hat{\Delta R}_t^i,\hat{\Delta x}_t^i)$.
We additionally supervise auxiliary distances $\hat{d}_{t+1,\mathrm{grip}}^i$ and $\hat{d}_{t+1,\mathrm{float}}^i$ from each scene object to the gripper and to its nearest floating object, encouraging the object features to encode geometric proximity for interaction modeling.

The predicted rigid motion is analytically applied to the object state. For each floating object, the rotation is applied to every 3D vector feature in the geometry latent, and the translation updates the object center:
\begin{equation}
\hat{\mu}_{t+1}^i=\hat{\Delta R}_t^i\circ\mu_t^i,
\qquad
\hat{c}_{t+1}^i=c_t^i+\hat{\Delta x}_t^i.
\end{equation}
The next object state is $\hat{s}_{t+1}^i=[\hat{\mu}_{t+1}^i;\hat{c}_{t+1}^i]$, while fixed objects remain unchanged and continue to provide static scene context. This structured update prevents the model from freely modifying object latents, preserving geometry by rolling out future states through object-level rigid motion. During multi-step prediction, ORD-WM recursively applies the predicted SE(3) updates in latent space, without repeated segmentation, point-cloud reconstruction, or object re-encoding.

We train the dynamics model with supervised object-wise SE(3) motion losses on floating objects and auxiliary distance losses on scene objects. Full loss definitions and training details are provided in Appendix~\ref{app:dynamics}.

\subsection{Model Predictive Control with ORD-WM}
\label{sec:method_mpc}

A learned world model can be used in several ways, including online planning, policy learning from imagined rollouts, and policy evaluation. In this work, we use ORD-WM for model predictive control, where the learned dynamics model is used directly to evaluate candidate action sequences.

At each control step, we construct the current scene state and proprioception, roll out future object states for candidate action sequences directly in latent space, and optimize the action sequence using the Cross-Entropy Method (CEM)~\citep{rubinstein1997optimization}. 
CEM iteratively samples candidate action sequences, evaluates them with the learned dynamics model and task cost, and refits the sampling distribution toward low-cost candidates. 
Only the first action of the optimized sequence is executed before the robot receives a new observation and replans, following a receding-horizon control scheme. 
The MPC objective measures the discrepancy between predicted scene states and desired target states over task-relevant components.

\begin{table}[t]
\centering
\small
\caption{Multi-step prediction comparison with PointWorld. Point-wise $\ell_2$ errors are reported for moving (mov.) and static (stat.) points at each horizon; Shape Consistency Chamfer Distance (CD) measures full-rollout drift from the initial object shape. All distances are in mm, and lower is better. PointWorld (ft) is fine-tuned on LIBERO. \textbf{Bold} marks the best result per metric.}
\label{tab:pointworld_comparison}
\setlength{\tabcolsep}{6pt}
\begin{tabular}{lccccccc}
\toprule
\multirow{2}{*}{Method}
& \multicolumn{2}{c}{1-step}
& \multicolumn{2}{c}{10-step}
& \multicolumn{2}{c}{Full traj.}
& \multirow{2}{*}{\makecell{Shape\\Cons. CD $\downarrow$}} \\
\cmidrule(lr){2-3} \cmidrule(lr){4-5} \cmidrule(lr){6-7}
& $\ell_2$ mov. & $\ell_2$ stat. & $\ell_2$ mov. & $\ell_2$ stat. & $\ell_2$ mov. & $\ell_2$ stat. & \\
\midrule
ORD-WM & \textbf{1.58} & \textbf{0.12} & \textbf{7.71} & \textbf{0.86} & \textbf{37.60} & \textbf{8.50} & \textbf{0.00} \\
PointWorld & 13.72 & 6.61 & 83.18 & 8.75 & 321.11 & 216.15 & 19.87 \\
PointWorld (ft) & 10.04 & 1.09 & 79.62 & 5.50 & 225.68 & 18.84 & 3.58 \\
\bottomrule
\end{tabular}
\par\vspace{3pt}
\begin{minipage}{\linewidth}
\footnotesize
For consistent point-based evaluation, ORD-WM's $\ell_2$ errors and consistency CD use initial-frame points transformed by the predicted cumulative SE(3) motions.
\end{minipage}
\end{table}

\begin{figure}[t]
    \vspace{-5pt}
    \centering
    \begin{subfigure}[b]{0.64\linewidth}
        \centering
        \includegraphics[width=\linewidth]{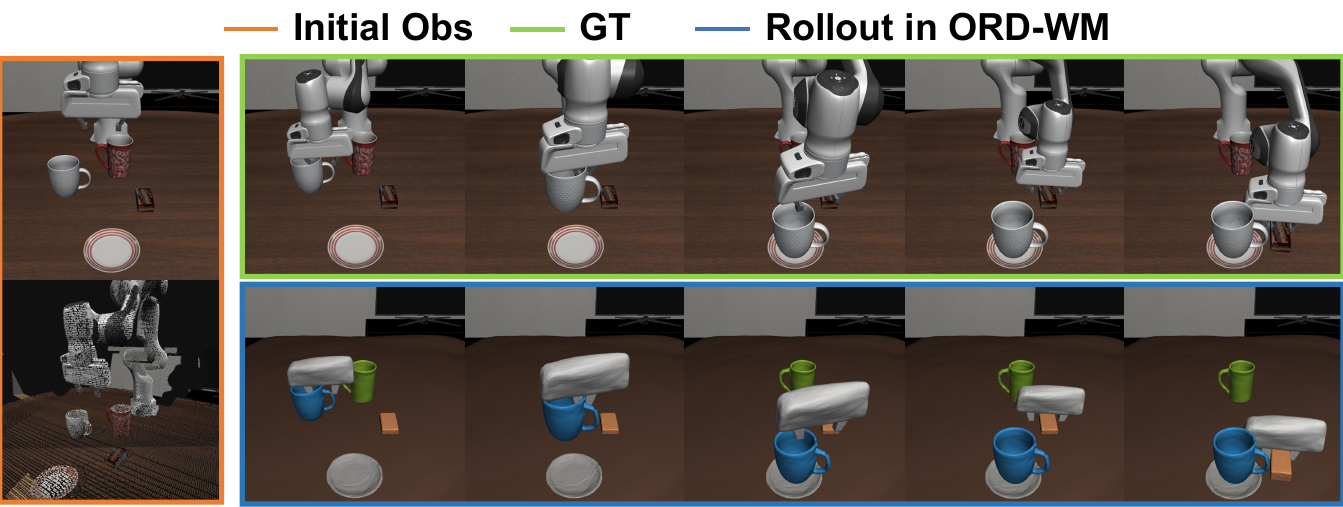}
        \label{fig:rollout_pred_3d}
    \end{subfigure}
    \hfill
    \begin{subfigure}[b]{0.345\linewidth}
        \centering
        \includegraphics[width=\linewidth]{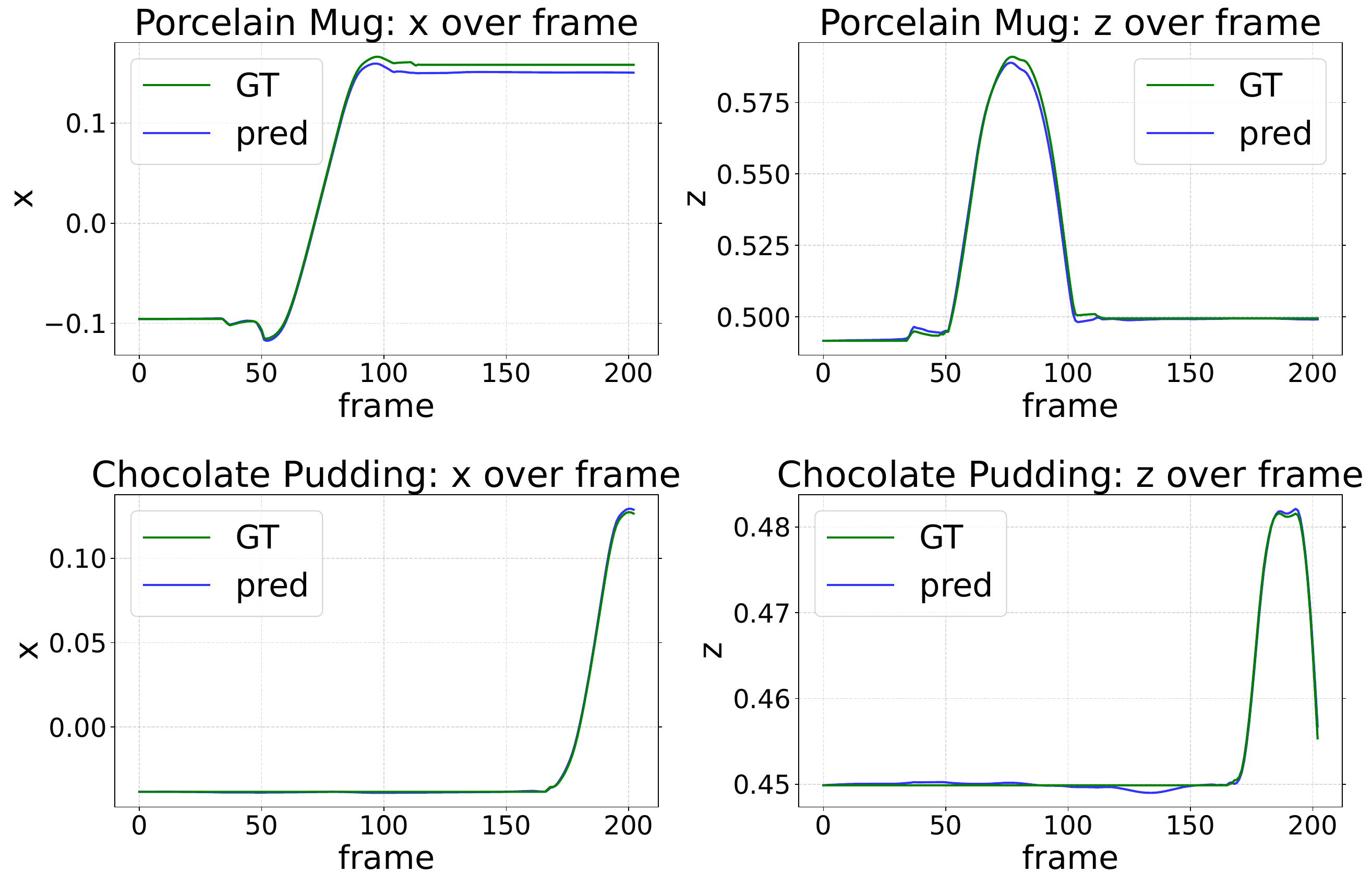}
        \label{fig:rollout_pred_time}
    \end{subfigure}

    \vspace{-0.6em}

    \begin{subfigure}[b]{0.49\linewidth}
        \centering
        \includegraphics[width=\linewidth]{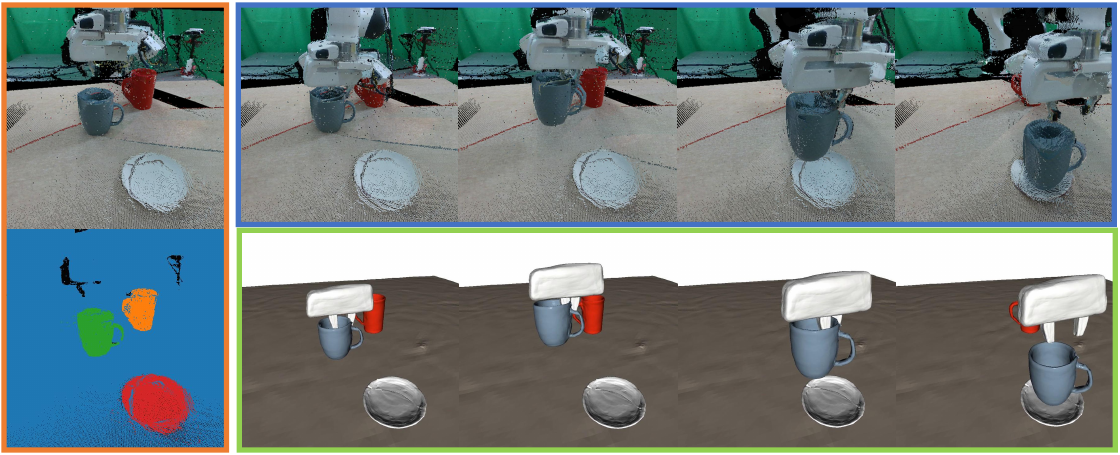}
        \label{fig:real_rollout_left}
    \end{subfigure}
    \hfill
    \begin{subfigure}[b]{0.49\linewidth}
        \centering
        \includegraphics[width=\linewidth]{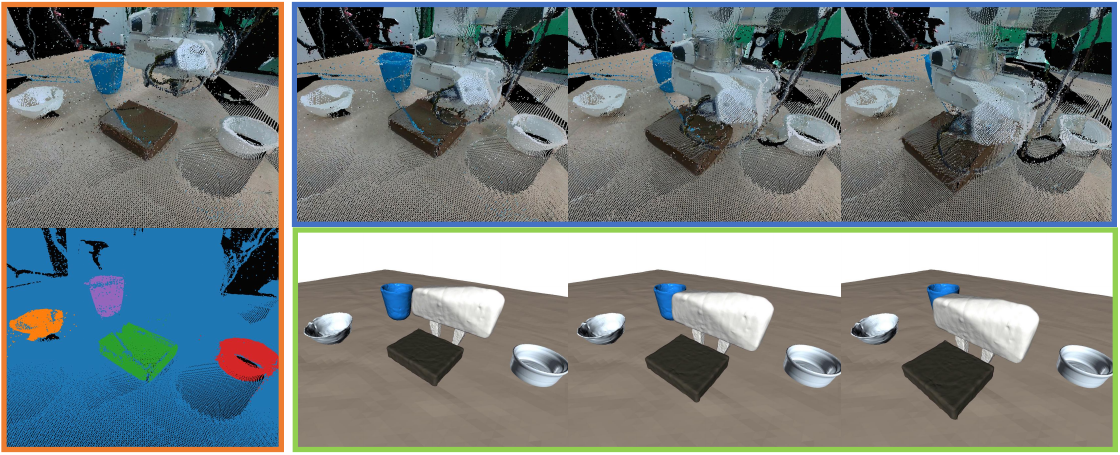}
        \label{fig:real_rollout_right}
    \end{subfigure}
    \vspace{-9pt}

    \caption{Qualitative rollout predictions. 
Top: simulation rollouts and object position trajectories compared with ground truth. 
Bottom: real-world rollouts from observed scenes.}
    \label{fig:rollout_prediction}
\end{figure}

\section{Experiments}

We conduct comprehensive experiments in simulation and the real world to answer the following questions:
(1)~Can ORD-WM accurately predict multi-step object motion while preserving geometry?
(2)~Which components enable stable geometry-preserving rollout?
(3)~Can ORD-WM support efficient closed-loop MPC?

\subsection{Experimental setup}

\noindent\textbf{Datasets.}
We train the SDF-VNVAE and its representation baselines using object assets from LIBERO~\citep{liu2023libero} scenes and RoboCasa~\citep{robocasa2024}. For world-model training, we use trajectories from the LIBERO benchmark, including LIBERO-Spatial, LIBERO-Object, LIBERO-Goal, and LIBERO-Long, each containing 10 distinct tasks, together with 1,000 additional object-pushing trajectories. We reserve 10\% of the trajectories for multi-step rollout evaluation.

\noindent\textbf{Real-world setup.}
Real-world experiments are conducted with a Franka Research 3 robot and three calibrated RealSense D435 cameras. Depth maps are estimated using FoundationStereo~\citep{wen2025stereo}, and object-centric states are constructed from the resulting RGB-D observations following the perception pipeline in Sec.~\ref{sec:method_obs}.

\subsection{Can ORD-WM accurately predict multi-step motion while preserving geometry?}
\label{exp:Q1}

We first evaluate whether ORD-WM can predict future object states while preserving geometry over multi-step rollouts. 
In simulation, we initialize ORD-WM with the initial scene state, roll it out using ground-truth action sequences from held-out trajectories, and compare the predictions with the corresponding ground truth at 1 step, 10 steps, and over the full trajectory.

We compare ORD-WM with the publicly released PointWorld checkpoint~\citep{huang2026pointworld} and a variant fine-tuned on LIBERO, denoted PointWorld (ft). Table~\ref{tab:pointworld_comparison} reports point-wise $\ell_2$ errors separately for moving and static points, together with Shape Consistency Chamfer Distance (CD). ORD-WM achieves lower prediction errors than both PointWorld variants for moving and static points across all evaluated horizons. Fine-tuning improves PointWorld's prediction accuracy and shape consistency, but geometric drift remains over full trajectories. In contrast, ORD-WM preserves object geometry throughout rollout by applying object-wise SE(3) updates.

Figure~\ref{fig:rollout_prediction} complements the quantitative results with representative rollout visualizations. 
In simulation, ORD-WM predicts object trajectories that closely follow ground truth over multiple steps, indicating that the latent rollout captures object-level motion beyond one-step transitions. 
In real-world scenes, the predicted rollouts are visually consistent with the corresponding observations, suggesting that the same rollout procedure transfers beyond simulation.
Additional rollout visualizations and generalization results are provided in Appendix~\ref{app:additional_rollout_results}.

\begin{table}[t]
\centering
\small
\caption{
Multi-step rollout prediction and efficiency. Surface CD is computed using points sampled from decoded object surfaces and evaluated at 1 step, 10 steps, and over full trajectories (avg. $\sim$170 steps). Shape Consistency CD measures full-rollout drift from the initial object shape. All CD values are in mm; rollout FPS uses batch sizes $B=1,8$. \textbf{Bold} marks the best reported result per metric.
}
\label{tab:rollout_prediction}
\setlength{\tabcolsep}{7pt}
\begin{tabular}{lcccccc}
\toprule
\multirow{2}{*}{Method}
& \multicolumn{3}{c}{Surface CD $\downarrow$}
& \multirow{2}{*}{\makecell{Shape\\Cons. CD $\downarrow$}}
& \multicolumn{2}{c}{Rollout FPS $\uparrow$} \\
\cmidrule(lr){2-4} \cmidrule(lr){6-7}
& 1-step & 10-step & Full traj. & & $B=1$ & $B=8$ \\
\midrule
ORD-WM
& \textbf{0.252} & \textbf{0.670} & \textbf{4.729} & $\mathbf{0}^{\dagger}$
& 784 & 6179 \\
w/o SE(3) update
& 0.620 & 2.588 & 131.600 & 4.683 & 902 & 6443\\
w/o VN and SE(3) update
& 8.501 & 9.763 & 52.401 & 34.611 & \textbf{1232} & \textbf{7439}\\
w/o VN
& 1.005 & 1.592 & 9.218 & $\mathbf{0}^{\dagger}$ & 162 & 124 \\
Decoupled Geo.+Pose
& 0.994 & 1.562 & 11.287 & $\mathbf{0}^{\dagger}$ & 891 & 6988 \\

\bottomrule

\end{tabular}
\par\vspace{3pt}
\begin{minipage}{\linewidth}
\footnotesize
$\dagger$ Geometry preserved by construction; numerical evaluation yields a residual below 0.05\,mm.
\end{minipage}
\vspace{-16pt}
\end{table}

\subsection{Which components enable stable geometry-preserving rollout?}
\label{exp:Q2}

We next analyze which structural components enable the stable rollouts observed in Sec.~\ref{exp:Q1}. 
At the core of ORD-WM are two key structures: a VN-based geometry latent that supports analytic object rotations, and a dynamics model that predicts object-wise SE(3) motion increments and applies them to the current object states in latent space.
To isolate their effects, we compare ORD-WM with four variants: two that directly predict the next object state and two that retain SE(3) updates without using VN representations.
The \emph{w/o SE(3) update} variant uses the same VN-based object state as ORD-WM but directly predicts the next object state in latent space, removing the analytic rigid update. 
The \emph{w/o VN and SE(3) update} variant replaces the VN-based geometry latent with a standard SDF-VAE latent and also predicts the next object state directly.
The \emph{w/o VN} baseline retains object-wise SE(3) prediction but uses a standard geometry latent. Since rotations cannot be applied directly to this latent, it transforms the object point cloud and re-encodes it at each rollout step.
The \emph{Decoupled Geo.+Pose} baseline stores persistent geometry and explicit pose separately, avoiding repeated geometry encoding. 
We evaluate these variants using Surface CD for prediction accuracy, Shape Consistency CD for geometric drift, and rollout FPS for computational efficiency.

The comparisons in Table~\ref{tab:rollout_prediction} show the value of SE(3) updates. For both VN-based and standard geometry representations, directly predicting future object states yields larger long-horizon errors than predicting and applying object-wise SE(3) motion increments. The direct-prediction variants also exhibit larger Shape Consistency CD, indicating that unconstrained latent updates can gradually alter the decoded geometry. In contrast, the variants with SE(3) updates preserve object shape by constraining state evolution to rigid motion. These results support the role of SE(3) updates in maintaining geometric consistency and improving multi-step prediction across both representations. Additional visualizations in Appendix~\ref{app:geometry_preservation_rollout} show how directly predicting future VN latents can accumulate geometric deformation over repeated rollout steps.

Among the variants that retain SE(3) updates, the choice of geometry representation further affects prediction accuracy and rollout efficiency. Compared with \emph{w/o VN}, ORD-WM achieves lower Surface CD at all horizons. It also achieves higher throughput by applying rotations directly to the VN latent, avoiding per-step point cloud re-encoding. The \emph{Decoupled Geo.+Pose} baseline also avoids re-encoding, but yields larger prediction errors across the evaluated horizons. A potential challenge of this factorization is reference-frame ambiguity: rotating the stored reference geometry and compensating in the explicit pose leaves the represented object unchanged. Unless a consistent reference-frame convention is enforced, equivalent physical states can therefore have different geometry--pose decompositions, potentially complicating dynamics learning. ORD-WM instead encodes geometry in its current orientation and updates it equivariantly, avoiding the need to select a separate canonical orientation.
Additional SDF-VNVAE analyses, including reconstruction metrics, reconstruction visualizations, latent interpolation, and latent-space clustering, are provided in Appendix~\ref{app:vnvae}, further supporting that the learned latent state captures reusable geometric structure.

\subsection{Can ORD-WM support efficient closed-loop MPC?}
\label{exp:Q3}

\begin{table}[t]
    \centering
    \small
    \caption{Real-world pushing MPC performance and planning efficiency. A trial is successful if the final object position error is below 2~cm. Pos. Error and Avg. Steps report the mean final position error and mean number of executed task steps, respectively. Planning Time is measured per MPC replanning step. \textbf{Bold} marks the best result per metric.
    }
    \label{tab:mpc_candidates}
    \setlength{\tabcolsep}{8pt}
    \begin{tabular}{@{}ccccc@{}}
    \toprule
    Candidates & Success Rate $\uparrow$ & Pos. Error $\downarrow$ & Avg. Steps $\downarrow$ & Planning Time $\downarrow$\\
    \midrule
    64  & 10\% & 3.72 cm & 367 & \textbf{55.4 ms} \\
    256  & 70\% & 1.64 cm & 171 & 115.7 ms \\
    1024 & \textbf{90\%} & \textbf{0.972 cm} & \textbf{85} & 269.9 ms \\
    \bottomrule
    \end{tabular}
\end{table}

\begin{figure}[t]
    \vspace{-2pt}
    \centering
    \includegraphics[width=0.99\textwidth]{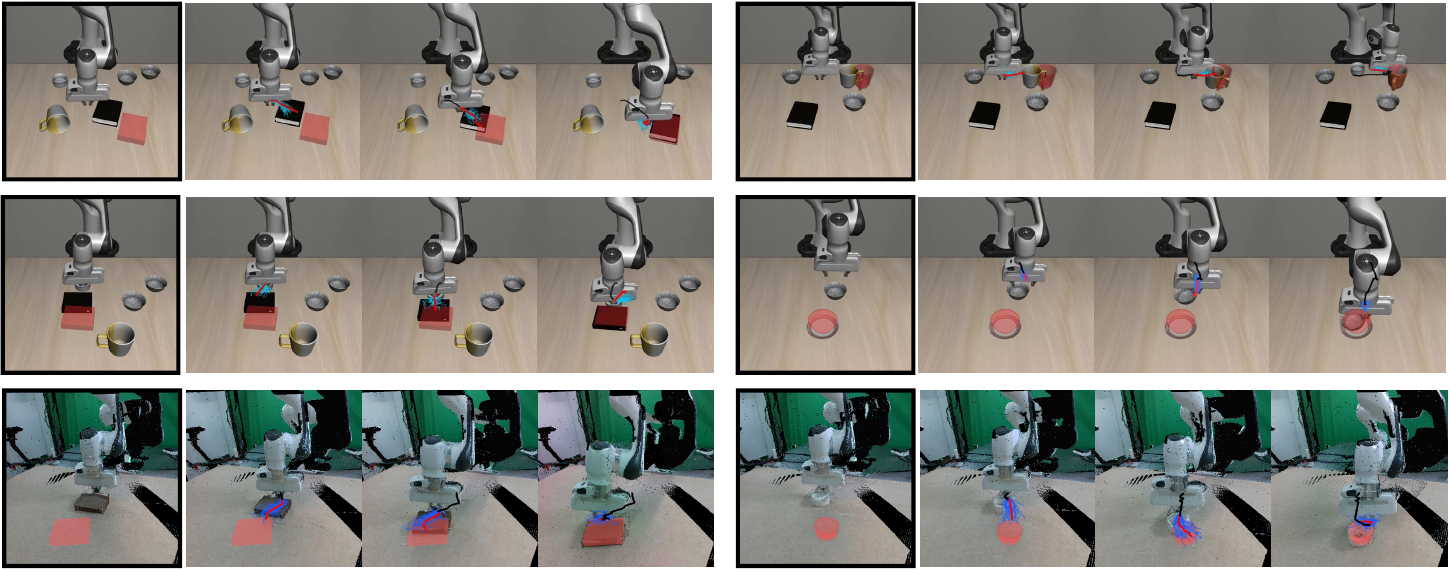}
    \caption{Closed-loop MPC with ORD-WM. The black-bordered panel shows the initial scene. At each control step, ORD-WM rolls out candidate trajectories in latent space (blue), selects the lowest-cost trajectory (red), and executes the first action in a receding-horizon manner; the accumulated executed trajectory is shown in black.}
    \label{fig:mpc}
    \vspace{-6pt}
\end{figure}

We finally evaluate whether ORD-WM can support closed-loop control through MPC. 
Following Sec.~\ref{sec:method_mpc}, the controller uses ORD-WM to roll out candidate action sequences in latent space and selects actions by minimizing the discrepancy between predicted scene states and desired target states over task-relevant components. We evaluate object pushing and pick-and-place tasks, measuring task performance and planning efficiency. For pick-and-place, we use a staged planning strategy inspired by V-JEPA 2~\citep{assran2025v}, sequentially performing grasping, lifting, and transport.

Table~\ref{tab:mpc_candidates} evaluates real-world pushing MPC performance under different candidate budgets, where each candidate is an 8-step action sequence, planning time is measured per MPC replanning step on a single NVIDIA RTX 3090 GPU, and success is defined as a final position error below 2~cm.
Increasing the number of candidates improves task performance: success rate rises from 10\% to 90\%, mean final position error decreases from 3.72~cm to 0.972~cm, and the average number of executed task steps drops from 367 to 85. 
Although larger budgets increase computation, planning remains below 0.3~s per replanning step with 1024 candidates on this hardware, indicating that ORD-WM supports repeated online replanning on the real robot.

Figure~\ref{fig:mpc} visualizes representative MPC trials in simulation and real-world scenes. In simulation, the selected actions move objects toward the target configurations over successive replanning steps. The real-world examples further show that ORD-WM can support closed-loop pushing from observed scenes beyond the simulation setting.

\section{Conclusion}
We proposed ORD-WM, a 3D world model that learns object-centric rigid dynamics for robotic manipulation.
By combining SDF-supervised, rotation-equivariant object representations with action-conditioned SE(3) motion prediction, ORD-WM evolves object states through analytic rigid transformations in latent space. This design preserves object geometry by construction and enables multi-step rollout without repeatedly reconstructing or re-encoding object point clouds.
Experiments demonstrate improved multi-step prediction accuracy, while ablations highlight the importance of rigid updates for geometry preservation and the benefits of equivariant representations for prediction accuracy and rollout efficiency. ORD-WM also supports efficient online replanning for closed-loop manipulation in both simulation and real-world experiments. Together, these results highlight the value of explicit rigid-body structure in learning world models for robotic manipulation.

\section{Limitations}
While ORD-WM introduces an object-centric SE(3)-structured latent rollout for preserving rigid-object geometry, its current formulation focuses on a restricted class of object dynamics. 
Specifically, scene objects are modeled as either floating or fixed rigid bodies, which limits the method's applicability to articulated, hinged, or deformable objects.
Future work could extend ORD-WM with additional motion types: articulated or hinged objects could be updated by predicting constrained transformations that respect their joint degrees of freedom, while deformable objects could use shape-changing latent dynamics instead of purely rigid SE(3) updates.

Even for rigid objects, ORD-WM targets low-dynamic, position-controlled manipulation. Mass, friction, and contact effects are learned implicitly from observed transitions, so the model cannot distinguish dynamics caused by different unobserved physical parameters under identical observed states and actions. Its SE(3) updates preserve object geometry but do not enforce contact constraints or non-penetration.

Finally, due to computational constraints, we have not yet evaluated ORD-WM on substantially larger datasets or more diverse object and task distributions. 
Scaling the proposed object-centric structured rollout to broader manipulation scenarios remains an important direction for future work.

\clearpage
\acknowledgments{
    This work was supported by National Natural Science Foundation of China 62673160, Shenzhen Fundamental Research ZDCY20250901095402003, Guangdong Provincial Natural Science Foundation 2026A1515011727 and the Open Fund of Innovation Center for Control Actuators. This work was also supported by the Analysis and Testing Center at Harbin Institute of Technology, Shenzhen.}

\bibliography{ref}

\appendix

\section{SDF-VNVAE Details and Analysis}
\label{app:vnvae}
\renewcommand{\theequation}{\Alph{section}.\arabic{equation}}
\renewcommand{\thefigure}{\Alph{section}.\arabic{figure}}
\renewcommand{\thetable}{\Alph{section}.\arabic{table}}

\makeatletter
\@addtoreset{equation}{section}
\@addtoreset{figure}{section}
\@addtoreset{table}{section}
\makeatother

\setcounter{equation}{0}
\setcounter{figure}{0}
\setcounter{table}{0}

This appendix provides additional details on the SDF-supervised vector-neuron variational autoencoder (SDF-VNVAE) used to construct the geometry-aware object states in ORD-WM. We describe the posterior parameterization, transformation properties, network architecture, training objective, training details, reconstruction accuracy, and additional qualitative analyses.

\subsection{Posterior Parameterization and Transformation Properties}
\label{app:vnvae_posterior}

For completeness, we provide the full posterior parameterization and clarify how the uncertainty term in the main text is computed. Given an object-centered point cloud $\bar{P}^i_t$, the VN encoder first outputs two sets of 3D vector features:
\begin{equation}
    (\mu^i_t, u^i_t) = \tilde{E}_\phi(\bar{P}^i_t),
    \qquad
    \mu^i_t \in \mathbb{R}^{K\times 3},
    \quad
    u^i_t \in \mathbb{R}^{K\times 3}.
\end{equation}
Here, $\mu^i_t$ contains the posterior mean vectors, and $u^i_t$ is an auxiliary vector-valued scale feature. The scalar uncertainty $\sigma^i_t \in \mathbb{R}^{K}$ used in the main text is computed from $u^i_t$ by taking the root mean square of each 3D auxiliary vector:
\begin{equation}
    \sigma^i_{t,k}
    =
    \operatorname{RMS}(u^i_{t,k})
    =
    \sqrt{
    \frac{1}{3}
    \left\|u^i_{t,k}\right\|_2^2
    + \epsilon
    },
    \qquad
    k=1,\ldots,K,
\end{equation}
where $\epsilon$ is a small constant for numerical stability. Thus, the notation
$(\mu^i_t,\sigma^i_t)=E_\phi(\bar{P}^i_t)$ in the main text should be understood as shorthand for this vector-to-scalar uncertainty parameterization.

The approximate posterior is factorized over the $K$ latent vector features:
\begin{equation}
    q_\phi(z^i_t \mid \bar{P}^i_t)
    =
    \prod_{k=1}^{K}
    \mathcal{N}
    \left(
    z^i_{t,k};
    \mu^i_{t,k},
    (\sigma^i_{t,k})^2 I_3
    \right),
    \qquad
    z^i_{t,k}\in\mathbb{R}^3.
\end{equation}
Therefore, each latent feature is a 3D Gaussian random vector with an isotropic covariance matrix. Importantly, $u^i_{t,k}$ is not used as an axis-wise standard deviation. Instead, it is converted into a single scalar uncertainty $\sigma^i_{t,k}$, which defines the isotropic covariance $(\sigma^i_{t,k})^2 I_3$.

This parameterization is compatible with SO(3)-equivariance. Since both $\mu^i_t$ and $u^i_t$ are produced by VN layers, they transform equivariantly under any rotation $R\in SO(3)$:
\begin{equation}
    \mu_\phi(R\circ \bar{P}^i_t)
    =
    R\circ \mu_\phi(\bar{P}^i_t),
    \qquad
    u_\phi(R\circ \bar{P}^i_t)
    =
    R\circ u_\phi(\bar{P}^i_t),
\end{equation}
where $R\circ X$ denotes applying the rotation $R$ to every point or 3D vector in $X$. The RMS operation then converts the equivariant auxiliary vector into a rotation-invariant scalar:
\begin{equation}
    \sigma_\phi(R\circ \bar{P}^i_t)
    =
    \sigma_\phi(\bar{P}^i_t),
\end{equation}
because
\begin{equation}
    \operatorname{RMS}(R u^i_{t,k})
    =
    \sqrt{
    \frac{1}{3}
    \left\|R u^i_{t,k}\right\|_2^2
    + \epsilon
    }
    =
    \sqrt{
    \frac{1}{3}
    \left\|u^i_{t,k}\right\|_2^2
    + \epsilon
    }
    =
    \operatorname{RMS}(u^i_{t,k}).
\end{equation}

As a result, the posterior distribution is equivariant as a distribution. For a single latent vector,
\begin{equation}
    z_k \sim \mathcal{N}(\mu_k, \sigma_k^2 I_3).
\end{equation}
After applying a rotation $R$, the transformed random vector satisfies
\begin{equation}
    R z_k
    \sim
    \mathcal{N}
    \left(
    R\mu_k,
    R(\sigma_k^2 I_3)R^\top
    \right)
    =
    \mathcal{N}
    \left(
    R\mu_k,
    \sigma_k^2 I_3
    \right).
\end{equation}
Thus, encoding the rotated input produces the same distribution as rotating samples drawn from the original posterior:
\begin{equation}
    z^i_t \sim q_\phi(z^i_t \mid \bar{P}^i_t)
    \quad \Longrightarrow \quad
    R\circ z^i_t \sim q_\phi(z^i_t \mid R\circ \bar{P}^i_t).
\end{equation}

During training, latent vectors are sampled using the reparameterization trick:
\begin{equation}
    z^i_{t,k}
    =
    \mu^i_{t,k}
    +
    \sigma^i_{t,k}\epsilon_k,
    \qquad
    \epsilon_k\sim\mathcal{N}(0,I_3).
\end{equation}
Since the standard Gaussian noise is rotation-invariant and $\sigma^i_{t,k}$ is a scalar, the sampled posterior remains equivariant in distribution. At test time, we use the posterior mean $\mu^i_t$ as the deterministic geometry latent, so the equivariance is exact at the level of the object state.

This design differs from using an axis-wise diagonal covariance for each 3D latent vector. If one directly interpreted a 3D vector as coordinate-wise standard deviations, the covariance would be
\begin{equation}
    \operatorname{diag}(\sigma_x^2,\sigma_y^2,\sigma_z^2).
\end{equation}
After rotation, this covariance would transform into
\begin{equation}
    R\operatorname{diag}(\sigma_x^2,\sigma_y^2,\sigma_z^2)R^\top,
\end{equation}
which is generally not diagonal in the original coordinate frame. Therefore, a coordinate-wise diagonal covariance would not preserve exact SO(3)-equivariance unless the full covariance transformation were explicitly modeled. By contrast, the isotropic covariance $(\sigma_k)^2 I_3$ remains invariant under rotation, while still allowing the uncertainty magnitude to be predicted from equivariant vector features.

The decoder predicts a signed distance value from the latent vectors and a query point:
\begin{equation}
    \hat{d}=D_\psi(z^i_t,x),
    \qquad
    x\in\mathbb{R}^3.
\end{equation}
It is designed to be invariant under a shared rotation of the latent representation and the query point:
\begin{equation}
    D_\psi(R\circ z^i_t, Rx)
    =
    D_\psi(z^i_t,x).
\end{equation}
Combining encoder equivariance with decoder invariance ensures that rotating the input object rotates the latent vectors, while the decoded signed distance field is preserved in the correspondingly rotated coordinate frame.

After training, ORD-WM uses the posterior mean as the deterministic object geometry latent and retains the object center separately:
\begin{equation}
    s^i_t=[\mu^i_t;c^i_t]\in\mathbb{R}^{(K+1)\times 3}.
\end{equation}
Given a predicted object-wise rigid motion $(\Delta R^i_t,\Delta x^i_t)$, the latent state is updated analytically as
\begin{equation}
    \hat{\mu}^i_{t+1}
    =
    \Delta R^i_t\circ \mu^i_t,
    \qquad
    \hat{c}^i_{t+1}
    =
    c^i_t+\Delta x^i_t.
\end{equation}
This transformation structure allows ORD-WM to apply object rotations directly to the geometry latent while using the object center to represent translation in the world frame.

\subsection{Network Architecture}
\label{app:vnvae_arch}

The SDF-VNVAE architecture follows the vector-neuron (VN) framework of \citet{deng2021vector}, which represents features as channels of three-dimensional vectors and defines SO(3)-equivariant neural operations such as VN linear layers, VN nonlinearities, pooling, normalization, and standardization. We adapt these operations to an SDF-supervised variational autoencoder for object geometry encoding. In the architecture tables, ``VNLinearLeakyReLU'' denotes a VN linear layer followed by a VN LeakyReLU nonlinearity.

All dimensions in VN layers are reported as vector-channel dimensions. A feature with \(C\) VN channels has shape \(\mathbb{R}^{C\times 3}\), corresponding to \(C\) three-dimensional vector features. This differs from a standard scalar MLP, where the hidden dimension counts scalar channels.

\paragraph{Input centering.}
For each object, let \(P=\{p_j\}_{j=1}^{N}\) denote its surface point cloud. Before encoding, the point cloud is centered by subtracting its centroid:
\begin{equation}
    \bar p_j = p_j - c,
    \qquad
    c=\frac{1}{N}\sum_{j=1}^{N}p_j .
\end{equation}
SDF query points are expressed in the same object-centered coordinate frame, and the decoder predicts signed distance values in this centered frame.

\paragraph{Encoder.}
The encoder is a VN-PointNet-style network without kNN graph construction. The centered point cloud is represented as one vector channel per point and processed by a shared VN trunk. Point-wise VN features are then fused with a pooled global VN feature, projected to a higher-dimensional vector representation, and mean-pooled over points to obtain an object-level representation. A VN latent head outputs \(64\) vector channels, which are split into posterior mean vectors and auxiliary scale vectors. In our implementation, the latent dimension is \(K=32\), with
\(\mu\in\mathbb{R}^{32\times 3}\) and auxiliary scale features in \(\mathbb{R}^{32\times 3}\). The encoder architecture is summarized in Table~\ref{tab:vnvae_encoder_arch}.

\begin{table}[h]
\centering
\caption{VN encoder architecture. Dimensions are reported as vector channels, and the batch dimension is omitted.}
\label{tab:vnvae_encoder_arch}
\begin{tabular}{lll}
\toprule
Component & Operation & Output shape \\
\midrule
Input & centered surface point cloud & \(2048 \times 3\) \\
VN format & coordinates as vector features & \(1 \times 3 \times 2048\) \\
Point embedding & VN layer with width \(32\) & \(32 \times 3 \times 2048\) \\
VN trunk & VN layers with widths \([32,64,64,64]\) & \(64 \times 3 \times 2048\) \\
Global feature & point-wise VN feature extraction and pooling & \(64 \times 3\) \\
Feature fusion & concatenate point and global features & \(128 \times 3 \times 2048\) \\
Projection & VN layers with widths \([256,1024]\) & \(1024 \times 3 \times 2048\) \\
Global pooling & mean pooling over points & \(1024 \times 3\) \\
Latent head & VN layers with widths \([512,512,512,64]\) & \(64 \times 3\) \\
Output split & mean and auxiliary scale vectors & \(32 \times 3,\;32 \times 3\) \\
\bottomrule
\end{tabular}
\end{table}

The first \(K=32\) output vectors are used as the posterior mean \(\mu\in\mathbb{R}^{32\times 3}\). The remaining \(32\) vectors are used as auxiliary scale features, from which the scalar uncertainty \(\sigma\in\mathbb{R}^{32}\) is computed as described in Sec.~\ref{app:vnvae_posterior}. Thus, the encoder notation used in the main text,
\begin{equation}
    (\mu,\sigma)=E_\phi(\bar P),
\end{equation}
is shorthand for a VN encoder that outputs vector-valued mean and auxiliary scale features, followed by a rotation-invariant mapping from the auxiliary scale features to isotropic uncertainty scalars.

\paragraph{Omitting kNN graph features.}
The original Vector Neurons work introduces VN variants of PointNet and DGCNN, and uses an input edge-convolution module in VN-PointNet to avoid the degeneracy that can occur when a shared VN-MLP is directly applied to single-vector point coordinates~\citep{deng2021vector}. In our final implementation, we omit kNN graph construction and directly process point coordinates with a VN-PointNet encoder. We found that this simplification slightly reduces reconstruction accuracy compared with the kNN-based variant, but improves efficiency and reduces sensitivity to the neighborhood structure. This trade-off is useful in our setting, where object point clouds are constructed from segmented RGB-D observations and can have non-uniform sampling density.

\paragraph{Decoder.}
The decoder predicts the signed distance value at an object-centered query point \(x\in\mathbb{R}^3\), conditioned on the latent vector set \(z\in\mathbb{R}^{32\times 3}\). For each query point, we concatenate \(x\) as an additional vector channel with the \(32\) latent vectors, yielding \(33\) input vector channels. The concatenated vector features are processed by VN layers, transformed by a VN standardization module, flattened into scalar features, and passed to an MLP that outputs a scalar SDF value. The decoder architecture is summarized in Table~\ref{tab:vnvae_decoder_arch}.

\begin{table}[h]
\centering
\caption{SDF decoder architecture. VN dimensions are reported as vector channels, while MLP dimensions are scalar channels. The batch and query dimensions are omitted.}
\label{tab:vnvae_decoder_arch}
\begin{tabular}{lll}
\toprule
Component & Operation & Output shape \\
\midrule
Latent input & latent vector set & \(32 \times 3\) \\
Query input & query point as vector channel & \(1 \times 3\) \\
Concatenation & latent and query vectors & \(33 \times 3\) \\
VN trunk & VN layers with widths \([128,128,128]\) & \(128 \times 3\) \\
VN standardization & canonical-frame projection & \(128 \times 3\) \\
Post-standardization VN layer & VN layer with width \(128\) & \(128 \times 3\) \\
Flatten & vector features to scalar features & \(384\) \\
Scalar MLP & MLP with widths \([256,256,256,1]\) & \(1\) \\
\bottomrule
\end{tabular}
\end{table}

The VN standardization module converts equivariant vector features into canonical-frame coordinates that are invariant to a shared input rotation. Since the latent vectors and query point rotate together under a shared rotation, the VN layers before standardization preserve equivariance, while the canonical-frame projection produces rotation-invariant scalar features for SDF prediction. Therefore, the decoder satisfies
\begin{equation}
    D_\psi(R\circ z, Rx)=D_\psi(z,x),
    \qquad R\in SO(3).
\end{equation}

\subsection{Training Objective}
\label{app:vnvae_loss}

The SDF-VNVAE is trained with an SDF reconstruction loss, an Eikonal regularization term, and a KL regularization term. Given a centered point cloud $\bar P$, the encoder parameterizes the posterior $q_\phi(z\mid \bar P)$, and the decoder predicts the signed distance value at an object-centered query point $x$:
\begin{equation}
    \hat d = D_\psi(z,x).
\end{equation}

\paragraph{SDF reconstruction loss.}
For query points $\{x_j\}_{j=1}^{Q}$ with ground-truth signed distance values $\{d_j\}_{j=1}^{Q}$, we use an $\ell_1$ reconstruction loss:
\begin{equation}
    \mathcal{L}_{\mathrm{sdf}}
    =
    \frac{1}{Q}
    \sum_{j=1}^{Q}
    \left|
    D_\psi(z,x_j)-d_j
    \right|.
\end{equation}

\paragraph{Eikonal regularization.}
To encourage the decoder to represent a valid signed distance field, we regularize the gradient norm with respect to the query point:
\begin{equation}
    \mathcal{L}_{\mathrm{eik}}
    =
    \frac{1}{2Q}
    \sum_{j=1}^{Q}
    \left|
    \left\|
    \nabla_x D_\psi(\operatorname{sg}(z),x_j)
    \right\|_2
    -
    1
    \right|,
\end{equation}
where $\operatorname{sg}(\cdot)$ denotes the stop-gradient operation. In other words, the latent variable is detached when computing the Eikonal term, so this regularization updates only the decoder parameters and does not backpropagate through the encoder. This prevents the Eikonal loss from shaping the posterior latent distribution directly, while still encouraging the decoder to produce a locally valid signed distance field around the queried points.

\paragraph{KL regularization.}
The approximate posterior is regularized toward an isotropic Gaussian prior,
\begin{equation}
    p(z)
    =
    \prod_{k=1}^{K}
    \mathcal{N}(0,I_3).
\end{equation}
Since each latent vector follows
\begin{equation}
    q_\phi(z_k\mid \bar P)
    =
    \mathcal{N}
    \left(
    \mu_k,\sigma_k^2 I_3
    \right),
\end{equation}
the KL term is
\begin{equation}
    \mathcal{L}_{\mathrm{KL}}
    =
    \frac{1}{2K}
    \sum_{k=1}^{K}
    \left(
    \|\mu_k\|_2^2
    +
    3\sigma_k^2
    -
    3
    -
    3\log\sigma_k^2
    \right).
\end{equation}
Here, $\sigma_k$ is the scalar uncertainty obtained from the auxiliary vector feature $u_k$ as described in Sec.~\ref{app:vnvae_posterior}.

The final training objective is
\begin{equation}
    \mathcal{L}_{\mathrm{VNVAE}}
    =
    \lambda_{\mathrm{sdf}}\mathcal{L}_{\mathrm{sdf}}
    +
    \lambda_{\mathrm{eik}}\mathcal{L}_{\mathrm{eik}}
    +
    \beta\mathcal{L}_{\mathrm{KL}}.
\end{equation}

\subsection{Training Details}
\label{app:vnvae_training}

We train the SDF-VNVAE using object assets from LIBERO and RoboCasa. For each object sample, surface points are used as the encoder input, and pre-generated SDF query points are used for decoder supervision. The query points are generated with a mixed sampling strategy and are shuffled and split into fixed-size chunks during training, rather than being re-sampled by category online. Here, batch size refers to the number of object samples per optimization step, while each object sample contains $N_p$ surface points and $Q$ SDF query points.

To account for imperfect RGB-D observations, we augment the input point clouds with partial or missing surfaces, nonuniform sampling, coordinate noise, outlier or background points, and simulated mask and camera perturbations. The object center is computed as the arithmetic mean of the observed points and can therefore vary with visibility and sampling density. These augmentations expose the encoder to the resulting variations in both the observed geometry and its centering.

During training, latent vectors are sampled from the posterior using the reparameterization trick. During evaluation and downstream ORD-WM training, we use the posterior mean as the deterministic geometry latent. The main training hyperparameters are summarized in Table~\ref{tab:vnvae_training}.

\begin{table}[h]
\centering
\caption{Training hyperparameters of the SDF-VNVAE.}
\label{tab:vnvae_training}
\begin{tabular}{ll}
\toprule
Hyperparameter & Value \\
\midrule
Surface points per object sample $N_p$ & $2048$ \\
SDF query points per object sample $Q$ & $2048$ \\
Query sampling strategy & pre-generated mixed samples \\
Sampling ratio & near / volume / far / shell = $0.75 / 0.12 / 0.08 / 0.05$ \\
Near-surface noise & $\sigma\sim\mathrm{Uniform}(0.001,0.05)$ \\
SDF truncation / clamping & none; raw SDF values are used \\
Optimizer & Adam \\
Encoder learning rate & $8\times 10^{-5}$ \\
Decoder learning rate & $8\times 10^{-5}$ \\
Weight decay & $10^{-6}$ \\
Scheduler & CosineAnnealingLR, $T_{\max}=200$ \\
Epochs & $200$ \\
Object samples per batch & $128$ \\
$\lambda_{\mathrm{sdf}}$ & $1.0$ \\
$\lambda_{\mathrm{eik}}$ & $0.004$ \\
$\beta$ & $0.002$ \\
KL annealing & none \\
\bottomrule
\end{tabular}
\end{table}

\subsection{Reconstruction Accuracy}
\label{app:vnvae_recon_metric}

We evaluate the reconstruction accuracy of the learned object representation on both seen and unseen object sets. LIBERO objects are used as the seen evaluation set, while OakInk2~\citep{zhan2024oakink2} objects are used as an unseen evaluation set. For each object, we encode the centered surface point cloud and use the posterior mean as the deterministic latent representation during evaluation. We evaluate reconstruction accuracy under the original object pose and under randomly sampled SO(3) rotations.

We compare three variants: a standard SDF-VAE, a standard SDF-VAE trained with pose augmentation, and our SDF-VNVAE. We report SDF $\ell_1$ error, Surface Chamfer Distance (CD), and Rotation Consistency CD, all in millimeters. The rotation consistency metric compares the reconstruction obtained from a rotated input with the rotated reconstruction obtained from the original input:
\begin{equation}
    \mathrm{CD}_{\mathrm{rot}}
    =
    \mathrm{CD}
    \left(
    \hat{\mathcal{S}}(R\circ \bar P),
    R\circ \hat{\mathcal{S}}(\bar P)
    \right),
    \qquad R\in SO(3),
\end{equation}
where $\hat{\mathcal{S}}(\bar P)$ denotes the reconstructed surface from the centered input point cloud $\bar P$. Lower Rotation Consistency CD indicates more consistent reconstruction under input rotations.

\begin{table}[t]
\centering
\caption{SDF reconstruction accuracy on seen and unseen object sets. LIBERO is used as the seen evaluation set, and OakInk2 is used as the unseen evaluation set. All distance-based metrics are reported in millimeters. Lower is better for all metrics. \textbf{Bold} indicates the best result for each metric.}
\label{tab:vnvae_recon}
\resizebox{\linewidth}{!}{
\begin{tabular}{llccccc}
\toprule
\multirow{2}{*}{Dataset} 
& \multirow{2}{*}{Method}
& \multicolumn{2}{c}{Original Pose}
& \multicolumn{2}{c}{Random SO(3) Pose}
& \multirow{2}{*}{\begin{tabular}[c]{@{}c@{}}Rotation\\Consistency CD\end{tabular}} \\
\cmidrule(lr){3-4}
\cmidrule(lr){5-6}
& & SDF L1 & Surface CD & SDF L1 & Surface CD & \\
\midrule
\multirow{3}{*}{LIBERO}
& SDF-VAE & 1.255 & 2.334 & 9.010 & 9.990 & 11.112 \\
& SDF-VAE + pose aug. & 1.938 & 3.457 & 1.913 & 3.406 & 3.023 \\
& SDF-VNVAE & \textbf{0.854} & \textbf{1.441} & \textbf{0.853} & \textbf{1.463} & \textbf{0.228} \\
\midrule
\multirow{3}{*}{OakInk2}
& SDF-VAE & 3.314 & 4.011 & 7.318 & 9.099 & 9.856 \\
& SDF-VAE + pose aug. & 2.176 & 2.960 & 2.171 & 3.692 & 4.673 \\
& SDF-VNVAE & \textbf{1.845} & \textbf{2.269} & \textbf{1.845} & \textbf{2.256} & \textbf{0.177} \\
\bottomrule
\end{tabular}
}
\end{table}

As shown in Table~\ref{tab:vnvae_recon}, SDF-VNVAE achieves the best reconstruction accuracy across both seen and unseen object sets. The standard SDF-VAE performs reasonably under the original pose but degrades substantially under random SO(3) rotations, indicating that the learned latent representation is sensitive to object orientation. Pose augmentation improves rotation robustness, but still yields higher reconstruction error and weaker rotation consistency than SDF-VNVAE. In contrast, SDF-VNVAE maintains nearly identical reconstruction accuracy under the original and randomly rotated poses, and achieves substantially lower Rotation Consistency CD. These results support that the vector-valued latent representation captures object geometry while transforming predictably under 3D rotations.

\subsection{Reconstruction Visualization}
\label{app:vnvae_recon_vis}

\begin{figure}[t]
\centering
\begin{subfigure}[t]{\linewidth}
    \centering
    \includegraphics[width=0.99\linewidth]{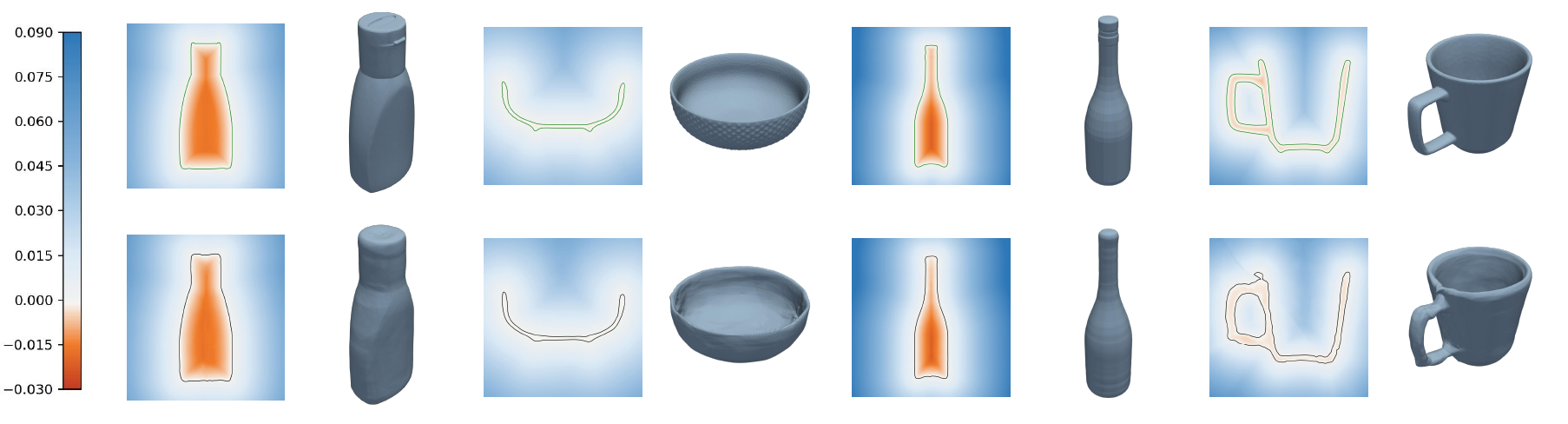}
    \caption{Seen LIBERO objects.}
    \label{fig:seen_obj_recon}
\end{subfigure}

\vspace{0.5em}

\begin{subfigure}[t]{\linewidth}
    \centering
    \includegraphics[width=0.99\linewidth]{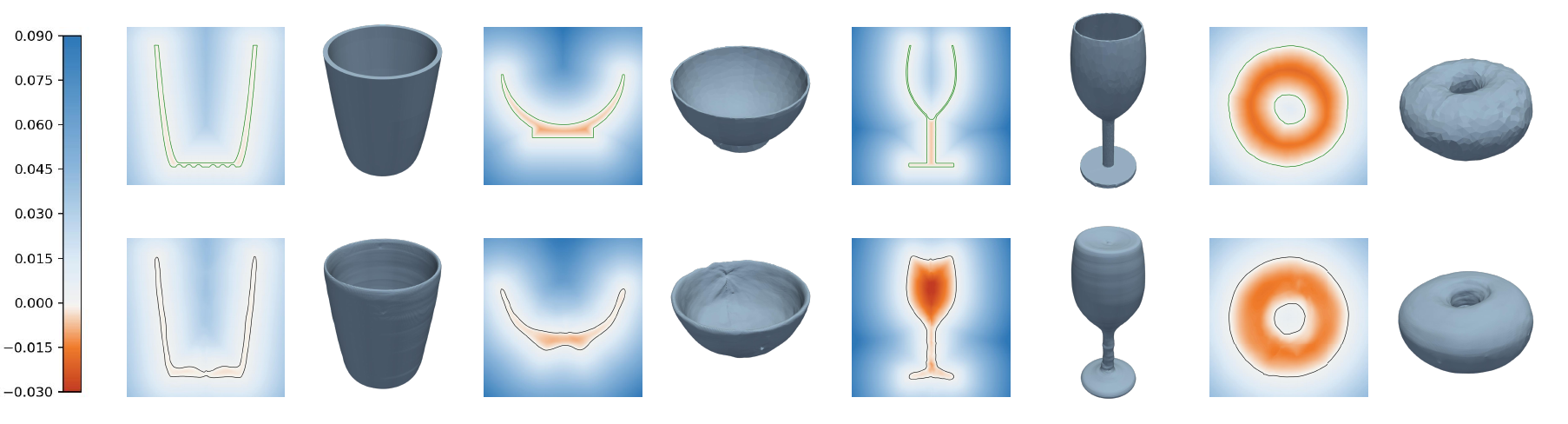}
    \caption{Unseen OakInk2 objects.}
    \label{fig:unseen_obj_recon}
\end{subfigure}
\caption{Reconstruction visualization of SDF-VNVAE on seen and unseen objects. For each example, we show a predicted SDF slice and the reconstructed zero-level surface.}
\label{fig:sdf_vnvae_recon}
\end{figure}

We further visualize the reconstruction results of SDF-VNVAE on both seen and unseen objects. Figure~\ref{fig:seen_obj_recon} shows examples from LIBERO objects, which correspond to the seen object set. Figure~\ref{fig:unseen_obj_recon} shows examples from OakInk2 objects, which are not used for training. For each example, we visualize a predicted SDF slice together with the reconstructed zero-level surface.

The qualitative results are consistent with the quantitative evaluation in Table~\ref{tab:vnvae_recon}. The SDF slices visualize the predicted distance structure around each object, while the zero-level surfaces provide a complementary view of the recovered three-dimensional shape. On seen LIBERO objects, SDF-VNVAE reconstructs diverse object geometries such as bottles, bowls, and mugs with accurate overall shape. On unseen OakInk2 objects, the model also produces reasonable reconstructions for novel shapes, suggesting that the learned vector-valued latent representation captures reusable geometric structure beyond the training object set. These visualizations further support that SDF-VNVAE learns geometry-aware object latents suitable for downstream object-centric rollout.

\subsection{Latent-Space Analysis}
\label{app:vnvae_latent}

We further analyze the learned latent space of SDF-VNVAE through latent interpolation and t-SNE visualization. These analyses qualitatively examine whether the learned vector-valued latents capture reusable geometric structure beyond point-wise reconstruction accuracy. Interpolation examines how decoded geometry varies between pairs of object representations, while t-SNE provides a complementary view of how representations of different objects are organized.

\begin{figure}[h]
\centering
\includegraphics[width=0.7\linewidth]{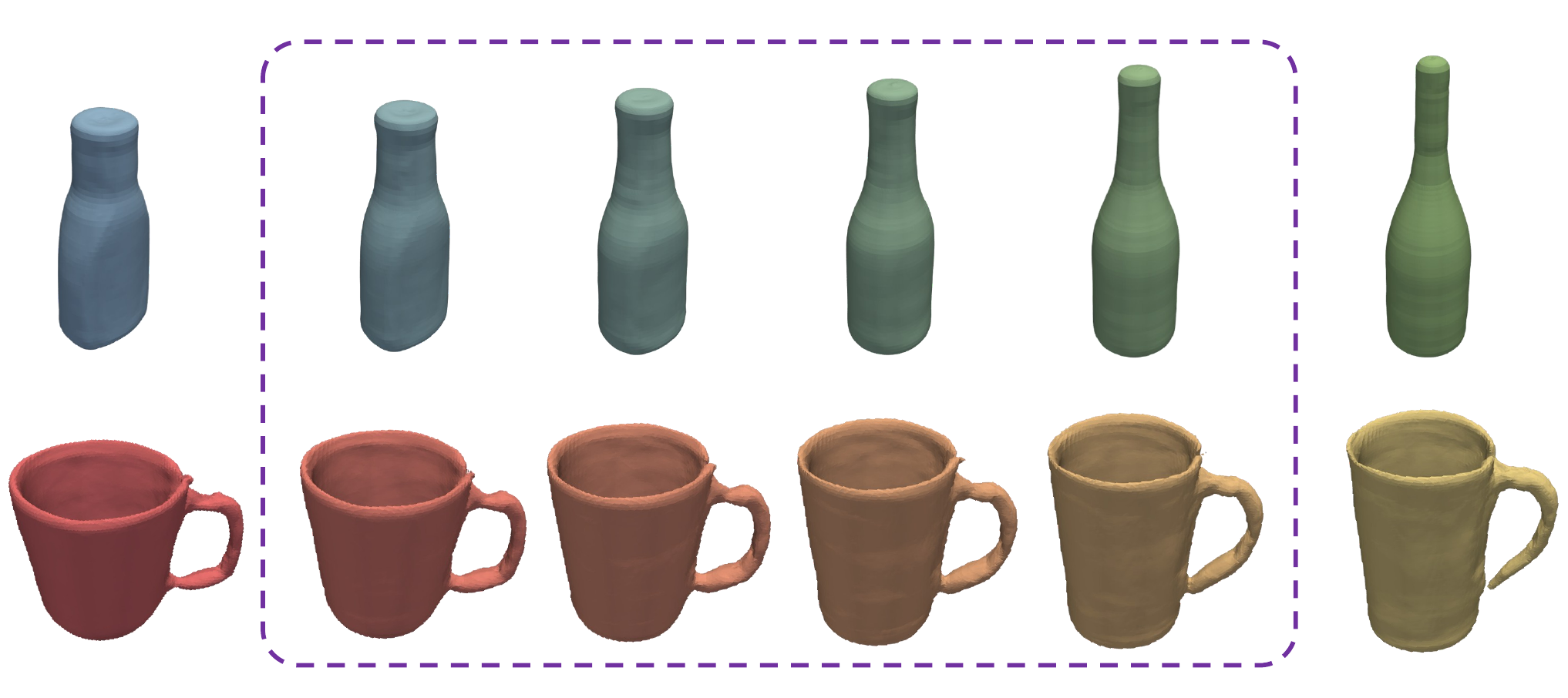}
\caption{Latent interpolation with SDF-VNVAE. The leftmost and rightmost shapes are reconstructed from endpoint object latents, while the shapes inside the dashed box are decoded from linearly interpolated latents. The smooth transitions indicate that the learned vector-valued latent space captures continuous object geometry.}
\label{fig:vnvae_interpolation}
\end{figure}

\paragraph{Latent interpolation.}
Given two object latents $\mu_a$ and $\mu_b$, we linearly interpolate between their posterior means:
\begin{equation}
    \mu(\alpha) = (1-\alpha)\mu_a + \alpha \mu_b,
    \qquad
    \alpha \in [0,1].
\end{equation}
The interpolated latents are decoded into SDF fields, and the zero-level surfaces are extracted for visualization. As shown in Fig.~\ref{fig:vnvae_interpolation}, the decoded shapes change smoothly between the two endpoint objects. For example, the bottle examples gradually vary in height and neck geometry, while the mug examples show smooth changes in body shape and handle geometry. This suggests that the learned latent space captures continuous geometric variations rather than only memorizing discrete object instances.

\begin{figure}[h]
\centering
\includegraphics[width=0.95\linewidth]{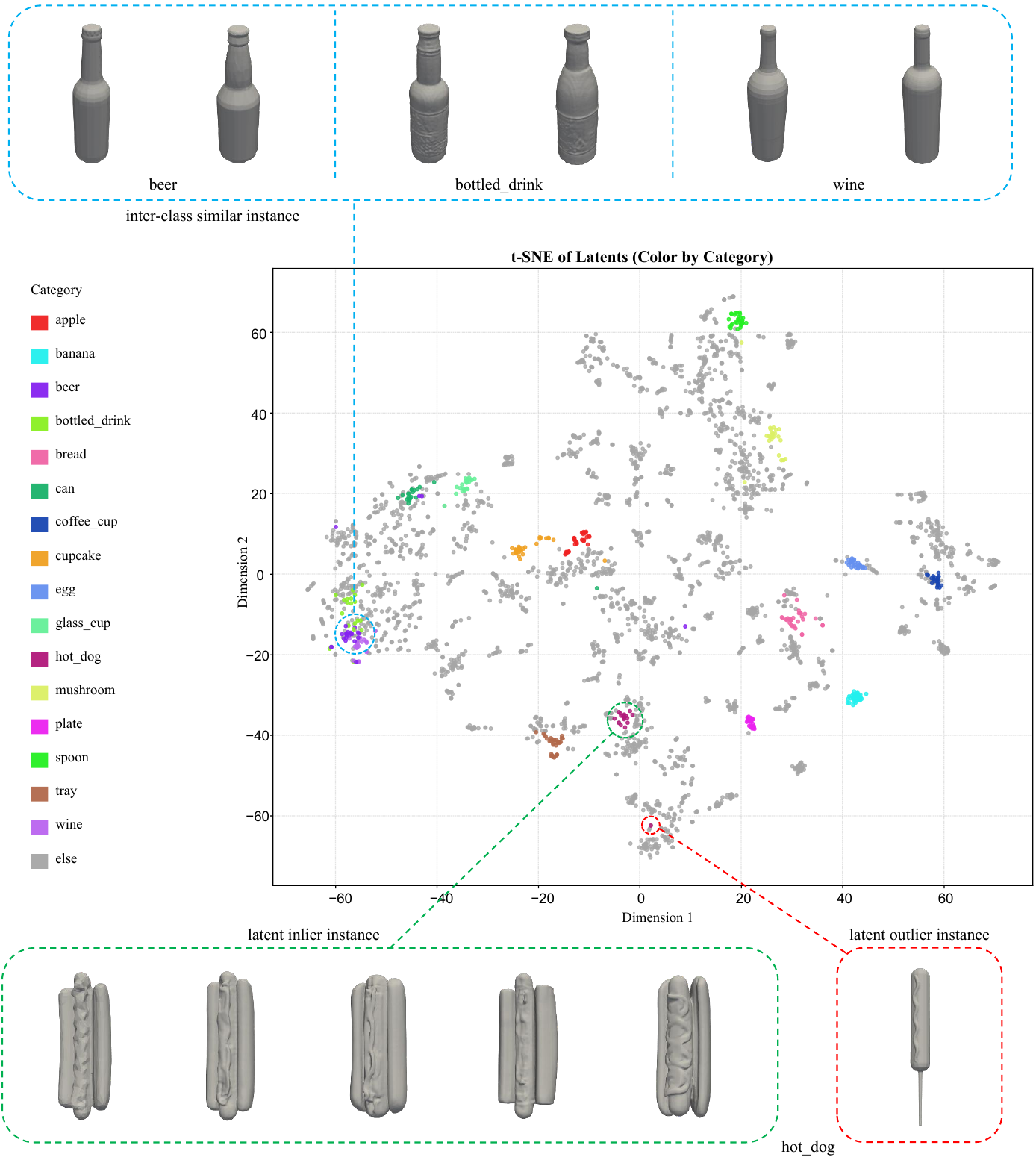}
\caption{t-SNE visualization of SDF-VNVAE object latents. Points are colored by object category. The latent space reflects geometric similarity: visually similar categories may appear close to each other, typical instances form compact local groups, and geometrically unusual instances can appear as outliers.}
\label{fig:vnvae_tsne}
\end{figure}

\paragraph{t-SNE visualization.}
We also visualize the learned object latents using t-SNE. Each point corresponds to an object latent, and colors indicate object categories. As shown in Fig.~\ref{fig:vnvae_tsne}, semantically and geometrically related objects tend to lie close to each other in the latent space. For example, bottle-like categories such as beer, bottled drink, and wine form nearby regions due to their similar elongated shapes. The visualization also reveals shape-level structure within a category: typical hot-dog instances form a compact group, while a visually distinct hot-dog instance appears as an outlier. These results suggest that the learned SDF-VNVAE latent space organizes objects according to geometric similarity, which is useful for representing object shape in downstream dynamics learning.

\section{Dynamics Model Details}
\label{app:dynamics}
This appendix provides additional implementation details of the dynamics model used in ORD-WM, including its network architecture, training objective, and training data construction.

\subsection{Network Architecture}
\label{app:dynamics_arch}
\begin{table}[h]
    \centering
    \caption{Architecture details of the dynamics model.}
    \label{tab:dynamics_arch}
    \renewcommand{\arraystretch}{1.08}
    \setlength{\tabcolsep}{5pt}
    \begin{tabular}{p{0.38\linewidth}p{0.54\linewidth}}
        \toprule
        Component & Configuration \\
        \midrule
        Backbone & Transformer encoder \\
        Object/Gripper token dimension & $99$ ($32 \times 3$ VN latent $+$ 3D center) \\
        Action token dimension & $9$ (6D rotation delta $+$ 3D translation delta) \\
        Object/Gripper token projection & $99 \rightarrow 512$ \\
        Action token projection & $9 \rightarrow 512$ \\
        Transformer hidden dimension & $512$ \\
        Transformer layers & $4$ \\
        Attention heads & $8$ \\
        Feed-forward dimension & $1024$ \\
        Activation & GELU \\
        Dropout & $0.0$ \\
        SE(3) motion head & $512 \rightarrow 9$ \\
        Gripper-distance head & $512 \rightarrow 512 \rightarrow 256 \rightarrow 1$ \\
        Floating-object-distance head & $512 \rightarrow 512 \rightarrow 256 \rightarrow 1$ \\
        \bottomrule
    \end{tabular}
\end{table}
The dynamics model is implemented as a Transformer encoder. Its input sequence is composed of four types of tokens: floating-object tokens, fixed-object tokens, gripper-part state tokens, and action tokens. Floating-object and fixed-object tokens encode the current object-centric scene state, where floating objects are allowed to move while fixed objects provide static scene context. The gripper-part state tokens describe the current poses and geometry of the gripper hand, left finger, and right finger. The action tokens encode the rigid pose changes of these same three gripper parts from time $t$ to $t+1$, with one token for each part. To preserve the correspondence between gripper parts and their actions, the three gripper-part tokens and the three action tokens are inserted into the sequence using a fixed part order, and each pair is assigned the same part-specific positional embedding. The Transformer produces interaction-aware features for scene objects, and the outputs at floating-object positions are used to predict object-wise SE(3) motion increments. The main architecture parameters are summarized in Table~\ref{tab:dynamics_arch}.

\subsection{Training Objective}
\label{app:dynamics_loss}
The dynamics model is trained with pose losses on floating objects and auxiliary distance losses on scene objects.

For each floating object, the predicted rotation and translation increments are composed with the current object pose:
\begin{equation}
    \hat{R}^i_{t+1} = \Delta \hat{R}^i_t R^i_t,
    \qquad
    \hat{c}^i_{t+1} = c^i_t + \Delta \hat{x}^i_t.
\end{equation}
The pose loss is
\begin{equation}
    \mathcal{L}_{\mathrm{pose}}
    =
    \lambda_{\mathrm{rot}}
    \frac{1}{|\mathcal{F}_t|}
    \sum_{i \in \mathcal{F}_t}
    \left\|
    \hat{R}^i_{t+1} - R^i_{t+1}
    \right\|^2_F
    +
    \lambda_{\mathrm{pos}}
    \frac{1}{|\mathcal{F}_t|}
    \sum_{i \in \mathcal{F}_t}
    \left\|
    \hat{c}^i_{t+1} - c^i_{t+1}
    \right\|^2_2.
\end{equation}

The two distance heads are supervised by masked MSE losses. The gripper-distance head predicts the nearest distance from each valid scene object to the gripper parts at $t+1$, and the floating-object-distance head predicts the nearest distance to floating objects at $t+1$:
\begin{equation}
\begin{aligned}
    \mathcal{L}_{\mathrm{dist}}
    =
    &\lambda_{\mathrm{grip}}
    \frac{1}{|\mathcal{M}_{\mathrm{grip}}|}
    \sum_{i \in \mathcal{M}_{\mathrm{grip}}}
    \left\|
    \hat{d}^i_{t+1,\mathrm{grip}} - d^i_{t+1,\mathrm{grip}}
    \right\|^2_2 \\
    & +
    \lambda_{\mathrm{float}}
    \frac{1}{|\mathcal{M}_{\mathrm{float}}|}
    \sum_{i \in \mathcal{M}_{\mathrm{float}}}
    \left\|
    \hat{d}^i_{t+1,\mathrm{float}} - d^i_{t+1,\mathrm{float}}
    \right\|^2_2 .
\end{aligned}
\end{equation}
Here, $\mathcal{M}_{\mathrm{grip}}$ and $\mathcal{M}_{\mathrm{float}}$ denote valid distance-supervision masks. Padding tokens and missing annotations are ignored.

The final objective is
\begin{equation}
    \mathcal{L}
    =
    \mathcal{L}_{\mathrm{pose}}
    +
    \mathcal{L}_{\mathrm{dist}}.
\end{equation}
In our experiments, all four loss weights are set to $1.0$.

\subsection{Dataset}
\label{app:dynamics_dataset}
We train the dynamics model using two types of trajectories: LIBERO manipulation trajectories and an additional object-pushing dataset. The LIBERO portion contains trajectories from LIBERO-Spatial, LIBERO-Object, LIBERO-Goal, and LIBERO-Long. Each suite contains 10 tasks, and each task provides 50 demonstrations, resulting in 2,000 LIBERO trajectories in total.

To increase the coverage of contact-rich rigid-object motion, we additionally collect 1,000 object-pushing trajectories in simulation. The pushing trajectories are collected fully automatically without human teleoperation. Each trajectory contains five consecutive pushing operations in the same scene. For each pushing operation, the data-collection script samples a movable target object and a planar pushing direction. Based on the target-object position and the sampled direction, the script automatically constructs a short waypoint sequence for the gripper: it first moves the gripper to a pre-contact pose behind the target object, then lowers the gripper to the pushing height, and finally moves the gripper along the sampled pushing direction.

The executed action sequence is generated by interpolating between these waypoints. To increase motion diversity, the execution speed is randomly sampled for each pushing operation, so trajectories contain pushes with different contact durations and motion magnitudes. After one pushing operation is completed, the resulting scene state is kept, and the next target object, pushing direction, and speed are sampled. Repeating this process five times within one trajectory produces diverse object displacements, contact configurations, and object-object interactions.

Both data sources are converted into the same center-aligned point-cloud trajectory format. For each frame, we store scene-object identities, object roles, object poses, gripper-part poses, and distance annotations. Scene objects are labeled as floating or fixed objects. Floating objects provide pose supervision for the SE(3) motion head, while fixed objects serve as static context. Object identities are matched with the pretrained SDF-VNVAE latent bank, and demonstrations without valid object matching or gripper information are filtered out.

For dynamics learning, each demonstration is converted into one-step transitions between consecutive frames. The current frame provides object and gripper-part states, the relative motions of the three gripper parts define the action tokens, and the next frame provides pose and distance supervision. We split each data source at the trajectory level into 90\% training and 10\% test data, ensuring that transitions from the same trajectory do not appear in both sets.

\subsection{Training Details}
\label{app:dynamics_training}

\begin{table}[t]
    \centering
    \caption{Training hyperparameters of the dynamics model.}
    \label{tab:dynamics_training}
    \renewcommand{\arraystretch}{1.08}
    \setlength{\tabcolsep}{6pt}
    \begin{tabular}{ll}
        \toprule
        Hyperparameter & Value \\
        \midrule
        Optimizer & AdamW \\
        Learning rate & $1 \times 10^{-4}$ \\
        Weight decay & $1 \times 10^{-6}$ \\
        Learning-rate schedule & Cosine decay \\
        Batch size & $256$ \\
        Epochs & $10000$ \\
        Context masking probability & $0.5$ \\
        \bottomrule
    \end{tabular}
\end{table}

The latents are obtained from the pretrained SDF-VNVAE latent bank, which includes posterior-mean representations of the augmented object observations described in Appendix~\ref{app:vnvae_training}. During training, we randomly sample one initial latent for each object identity. This exposes the dynamics model to observation-induced variations in the representation of the same object and reduces overfitting to a single latent realization.
For each training step, object tokens and gripper-part tokens are constructed from the current object states and gripper states. Action tokens are constructed from the relative rigid motions of the three gripper parts between two consecutive frames. The model predicts SE(3) motion increments for floating objects, which are composed with the current object poses before computing the pose losses. The auxiliary distance heads are supervised with next-step nearest-distance targets.

We use masks to handle variable-size scenes and missing annotations. The token mask removes padded tokens. The floating-object mask restricts pose losses to valid floating objects. The distance masks remove objects without valid gripper-distance or floating-object-distance annotations.
We additionally apply group-wise context masking during training. Object groups marked as mask-eligible are randomly removed with probability $0.5$, and the corresponding token masks and loss masks are updated accordingly. This serves as a data augmentation that reduces reliance on non-critical static context objects.

The main training hyperparameters are summarized in Table~\ref{tab:dynamics_training}. We train the model with AdamW and a cosine learning-rate schedule. The materialized transitions are preloaded onto GPU memory before training to reduce HDF5 reading overhead and host-to-device transfer cost.

\raggedbottom
\section{Additional Rollout Results and Analysis}
\label{app:additional_rollout_results}

This appendix provides supplementary rollout results and analyses that complement the main experimental evaluation. We include additional multi-step rollout visualizations on held-out trajectories, counterfactual rollouts under different future actions and object substitutions, and an analysis of geometry preservation during latent-space rollout.

\subsection{Additional Multi-Step Rollout Results}
\label{app:additional_multistep_rollout}

We provide additional multi-step rollout visualizations on held-out test trajectories to complement the qualitative results in Fig.~\ref{fig:rollout_prediction}. These examples cover different initial object configurations and interaction patterns, and show both the predicted object rollouts and the corresponding object position trajectories compared with ground truth.

\begin{figure}[H]
\centering
\setlength{\tabcolsep}{0pt}
\begin{tabular}{@{}c@{}}
\includegraphics[width=0.9\linewidth]{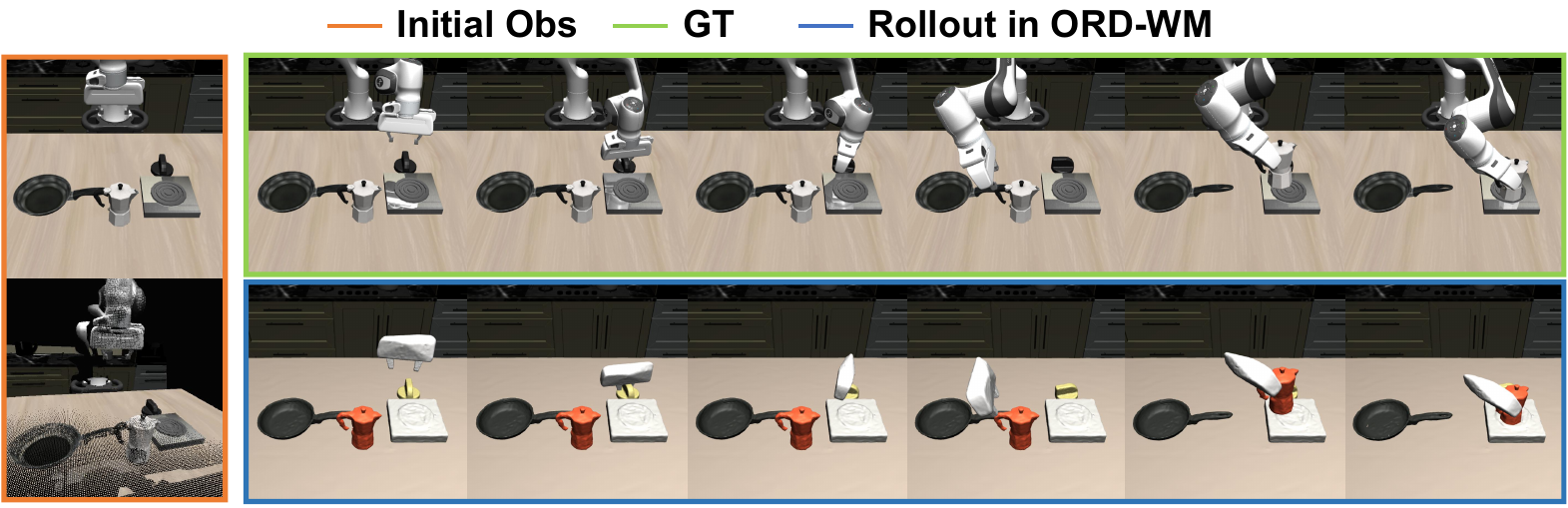} \\
\includegraphics[width=0.95\linewidth]{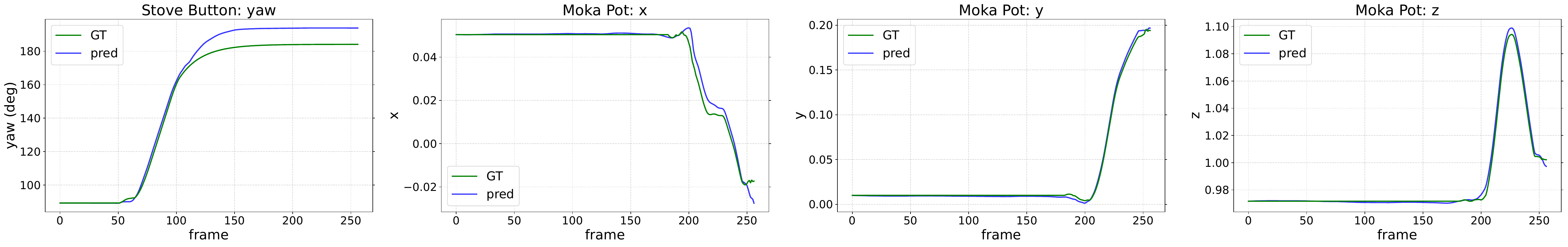}
\end{tabular}
\caption{Additional multi-step rollout results on held-out trajectories, part 1.}
\label{fig:additional_rollout_test_part1}
\end{figure}

\begin{figure}[H]
\centering
\setlength{\tabcolsep}{0pt}
\begin{tabular}{@{}c@{}}
\includegraphics[width=0.9\linewidth]{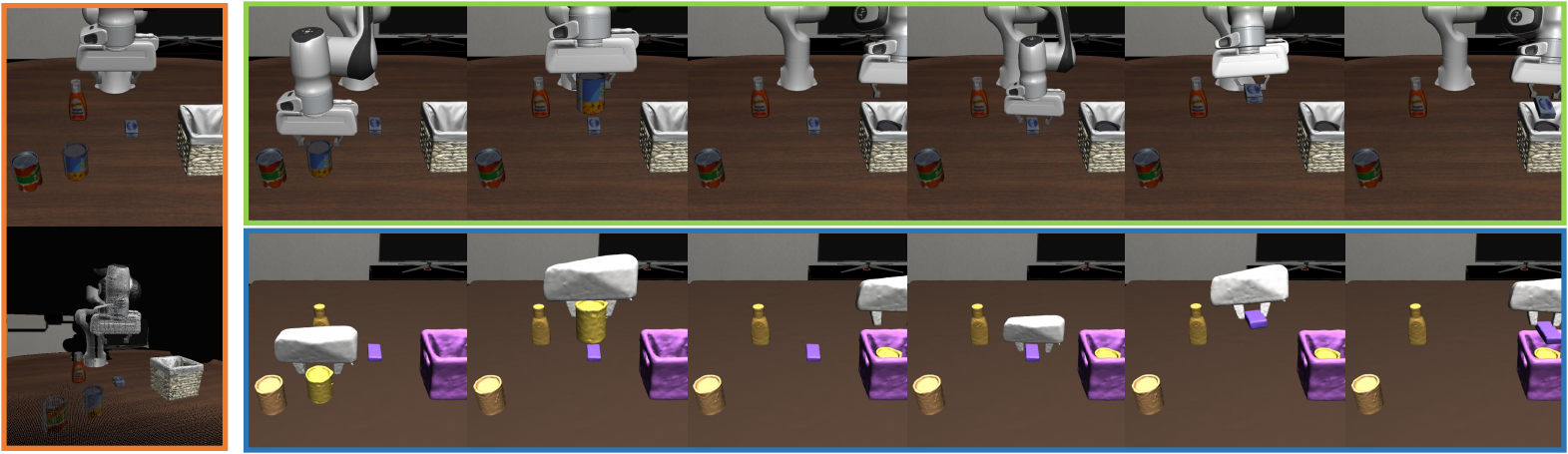} \\
\includegraphics[width=0.75\linewidth]{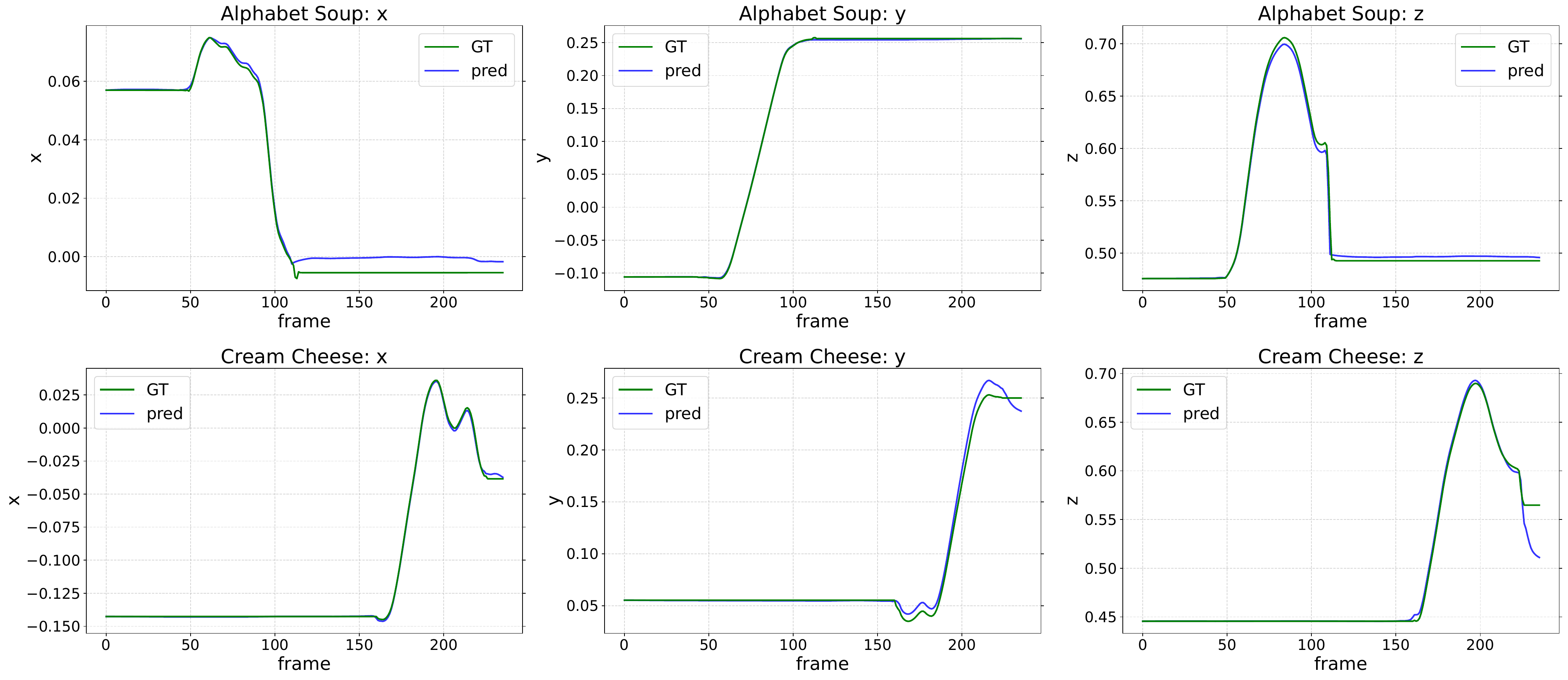} \\[0.25em]
\includegraphics[width=0.9\linewidth]{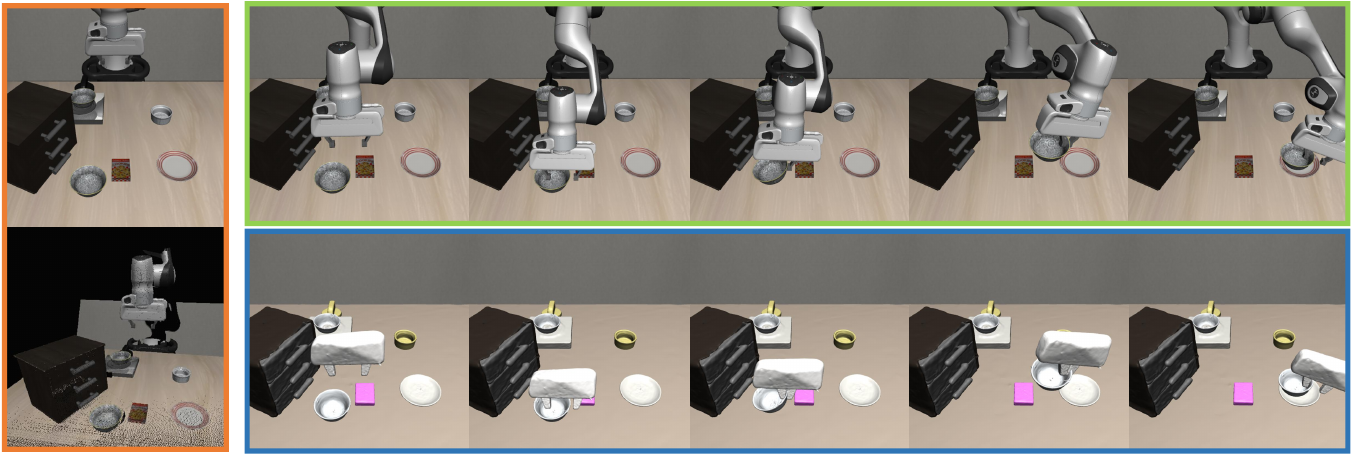} \\
\includegraphics[width=0.75\linewidth]{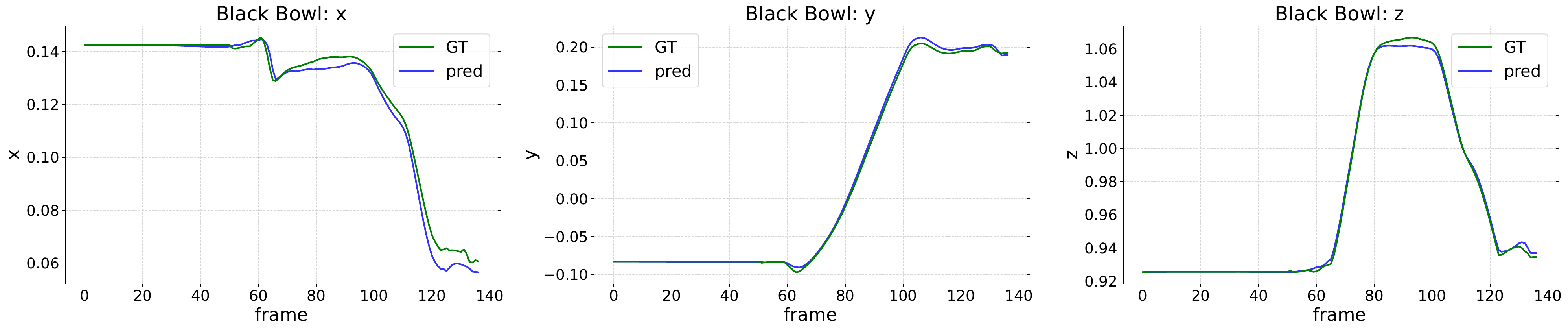} \\[0.25em]
\includegraphics[width=0.9\linewidth]{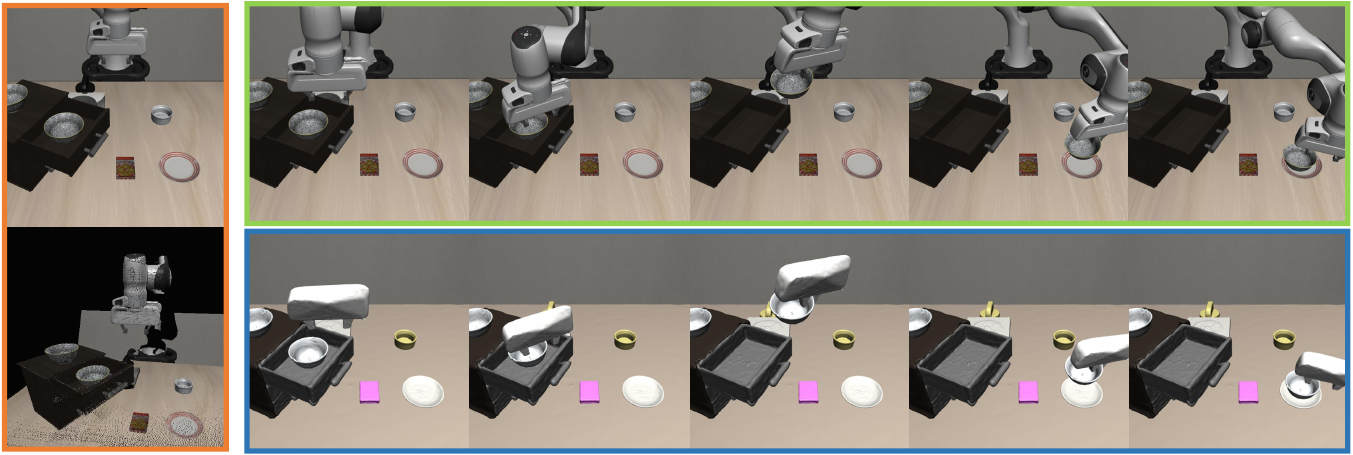} \\
\includegraphics[width=0.75\linewidth]{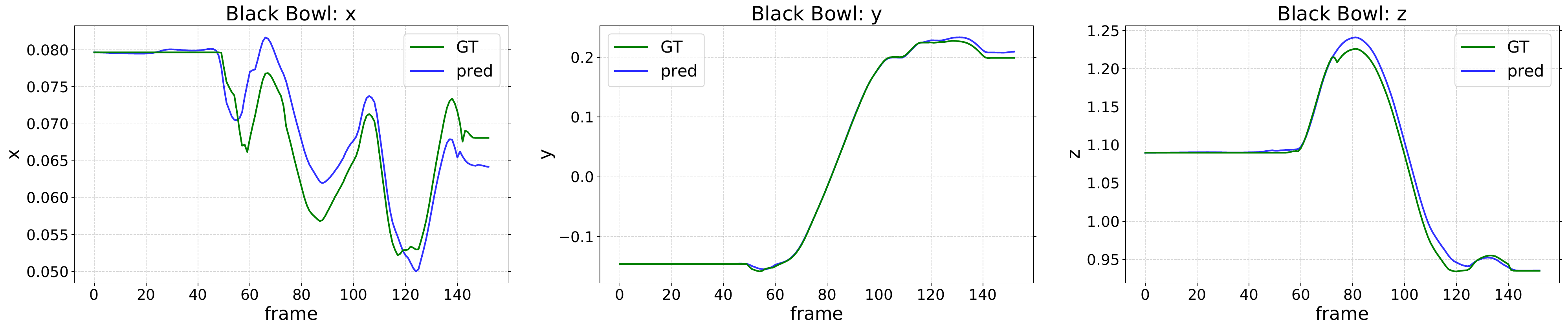}
\end{tabular}
\caption{Additional multi-step rollout results on held-out trajectories, part 2.}
\label{fig:additional_rollout_test_part2}
\end{figure}

\begin{figure}[H]
\centering
\setlength{\tabcolsep}{0pt}
\begin{tabular}{@{}c@{}}
\includegraphics[width=0.9\linewidth]{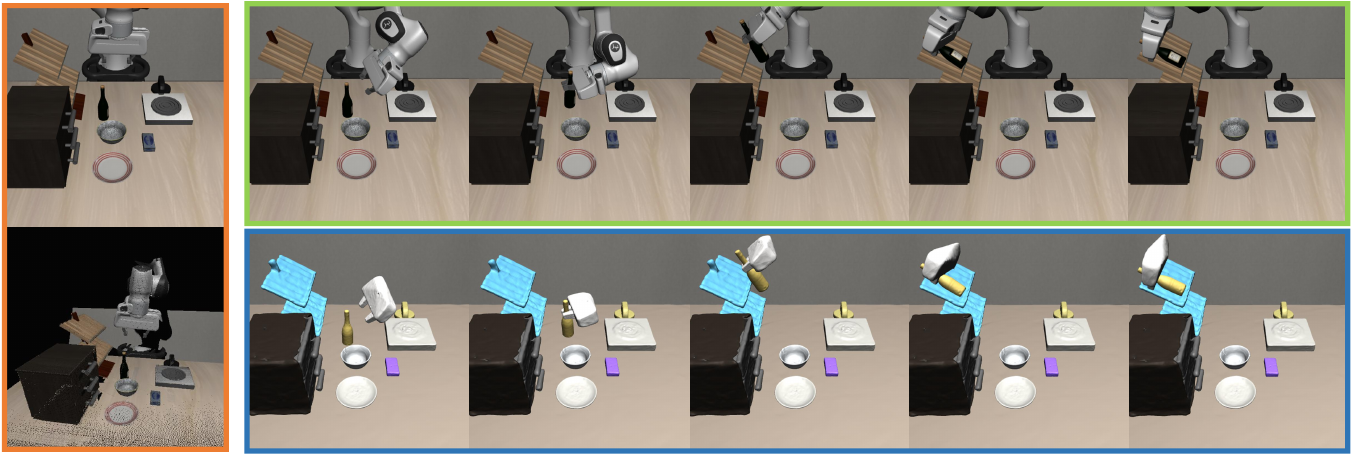} \\
\includegraphics[width=0.85\linewidth]{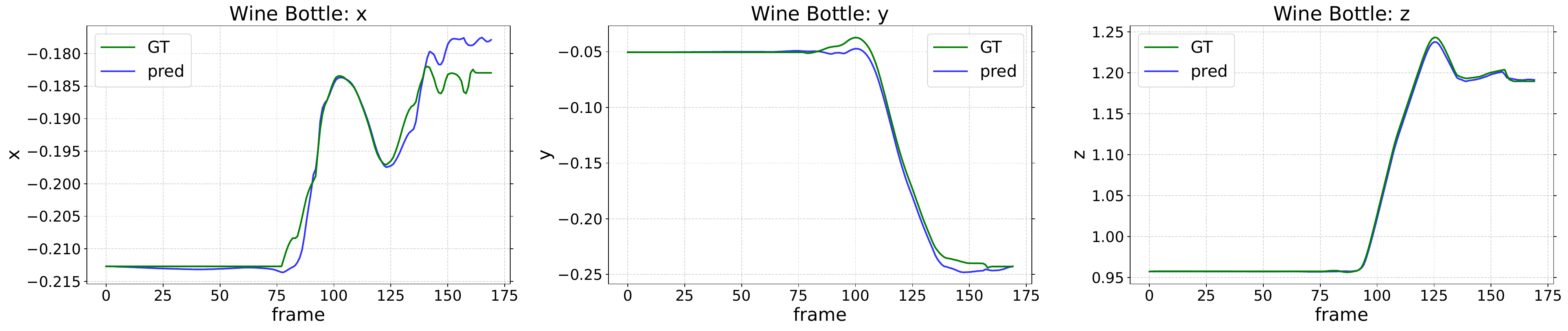} \\[0.25em]
\includegraphics[width=0.9\linewidth]{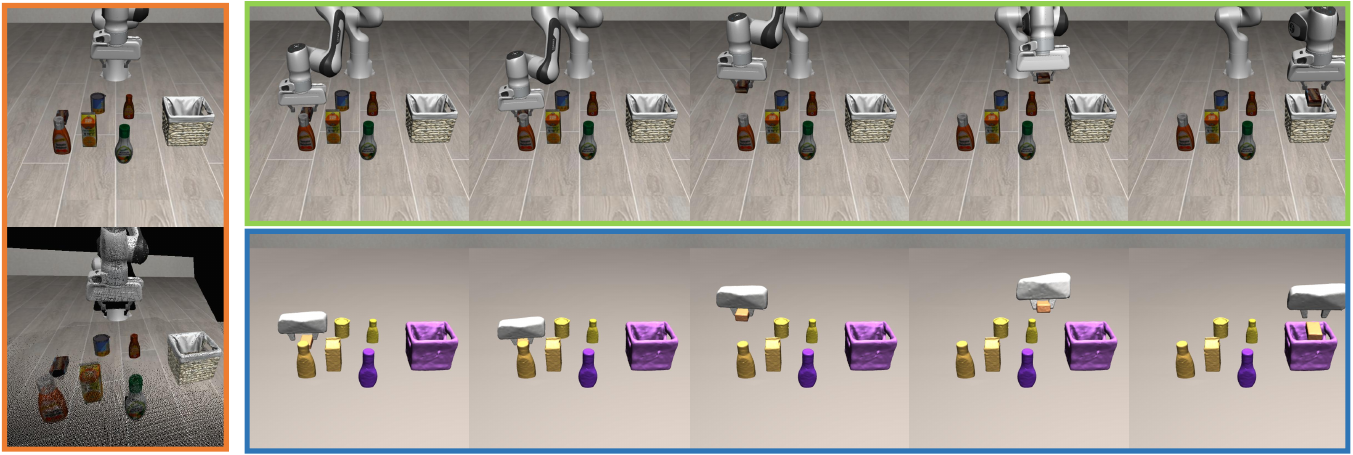} \\
\includegraphics[width=0.85\linewidth]{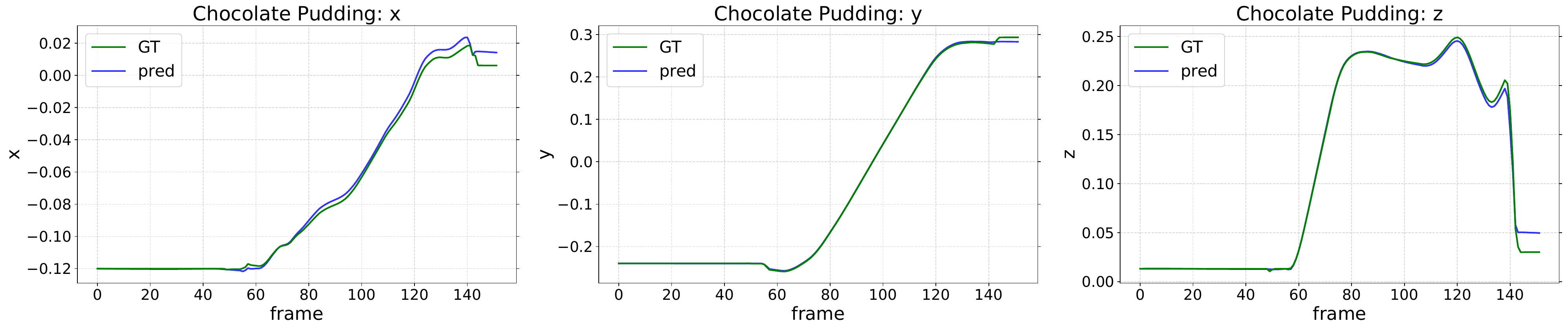} \\[0.25em]
\includegraphics[width=0.9\linewidth]{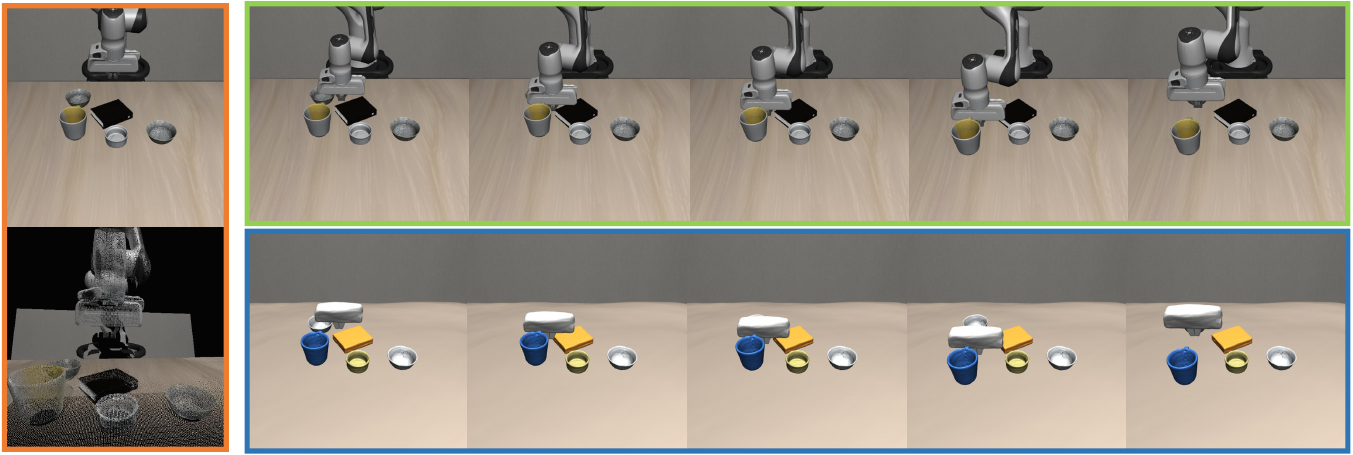} \\
\includegraphics[width=0.85\linewidth]{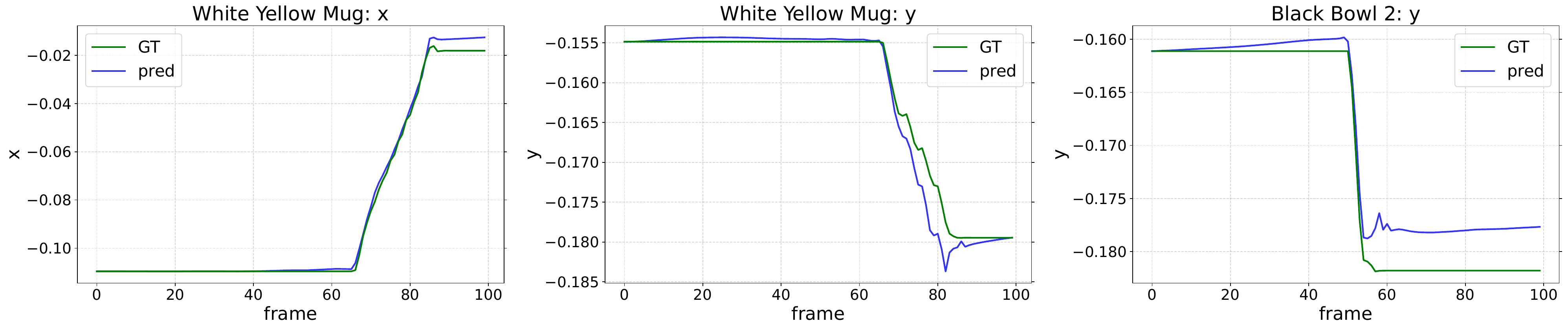}
\end{tabular}
\caption{Additional multi-step rollout results on held-out trajectories, part 3.}
\label{fig:additional_rollout_test_part3}
\end{figure}

\begin{figure}[H]
\centering
\setlength{\tabcolsep}{0pt}
\begin{tabular}{@{}c@{}}
\includegraphics[width=0.9\linewidth]{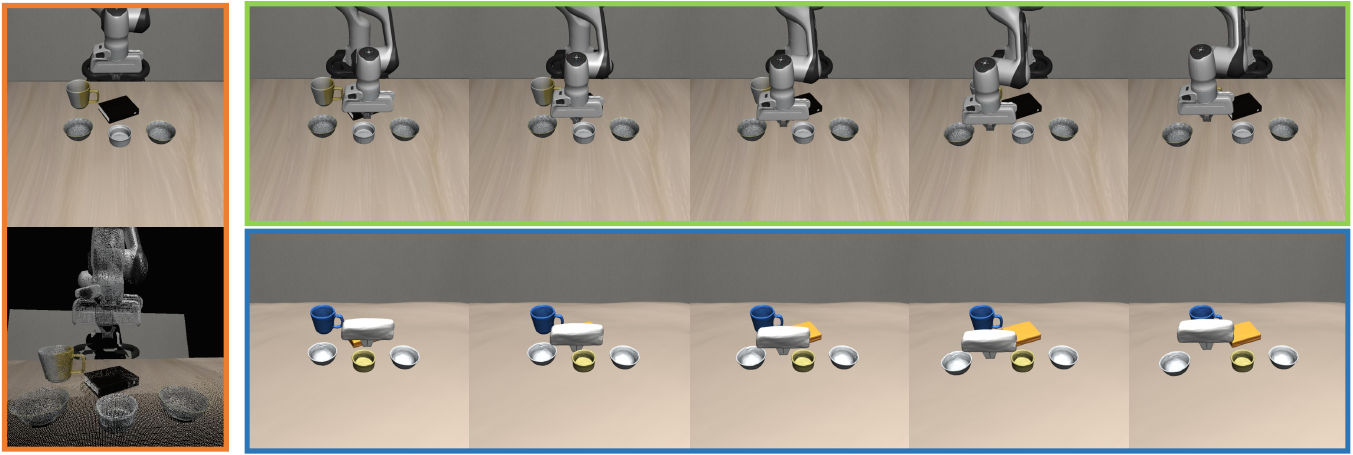} \\
\includegraphics[width=0.5\linewidth]{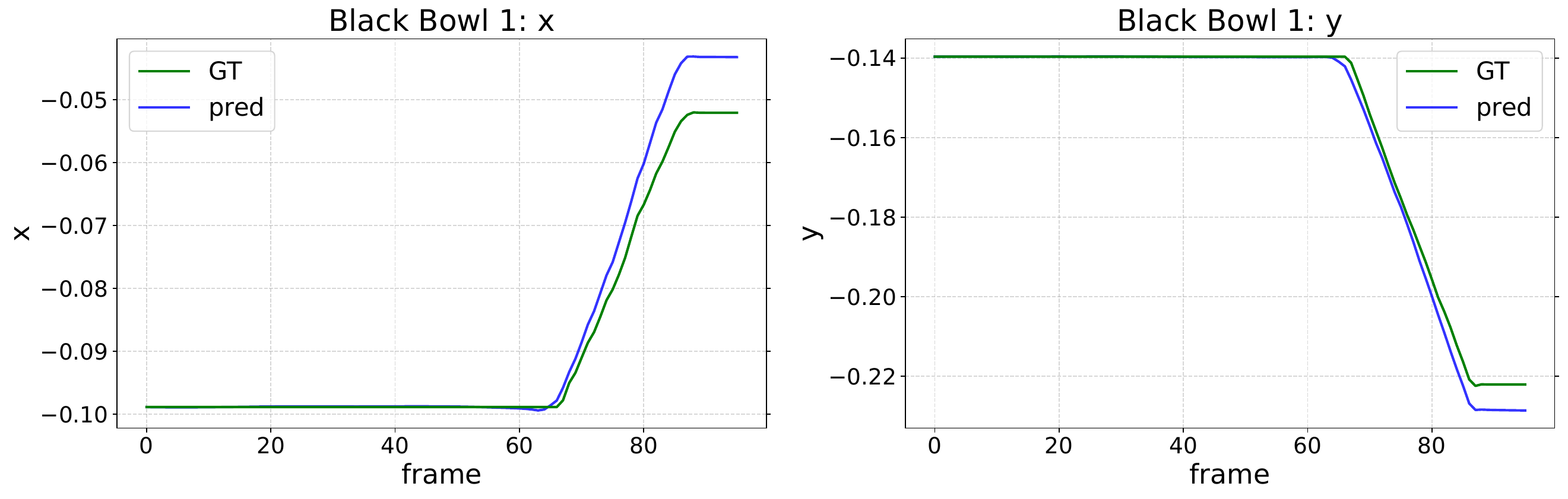}
\end{tabular}
\caption{Additional multi-step rollout results on held-out trajectories, part 4.}
\label{fig:additional_rollout_test_part4}
\end{figure}

As shown in Figs.~\ref{fig:additional_rollout_test_part1}--\ref{fig:additional_rollout_test_part4}, ORD-WM generally follows the ground-truth object motion across the held-out trajectories. In most examples, the predicted visual rollouts remain consistent with the ground-truth evolution, and the position trajectories closely track the corresponding ground-truth curves. These results further indicate that the learned dynamics model can predict multi-step object motion beyond one-step transitions while maintaining stable latent-space rollout behavior.

\subsection{Counterfactual Action Rollouts}
\label{app:rollout_action_conditioned}

\begin{figure}[H]
\centering
\includegraphics[width=0.82\linewidth]{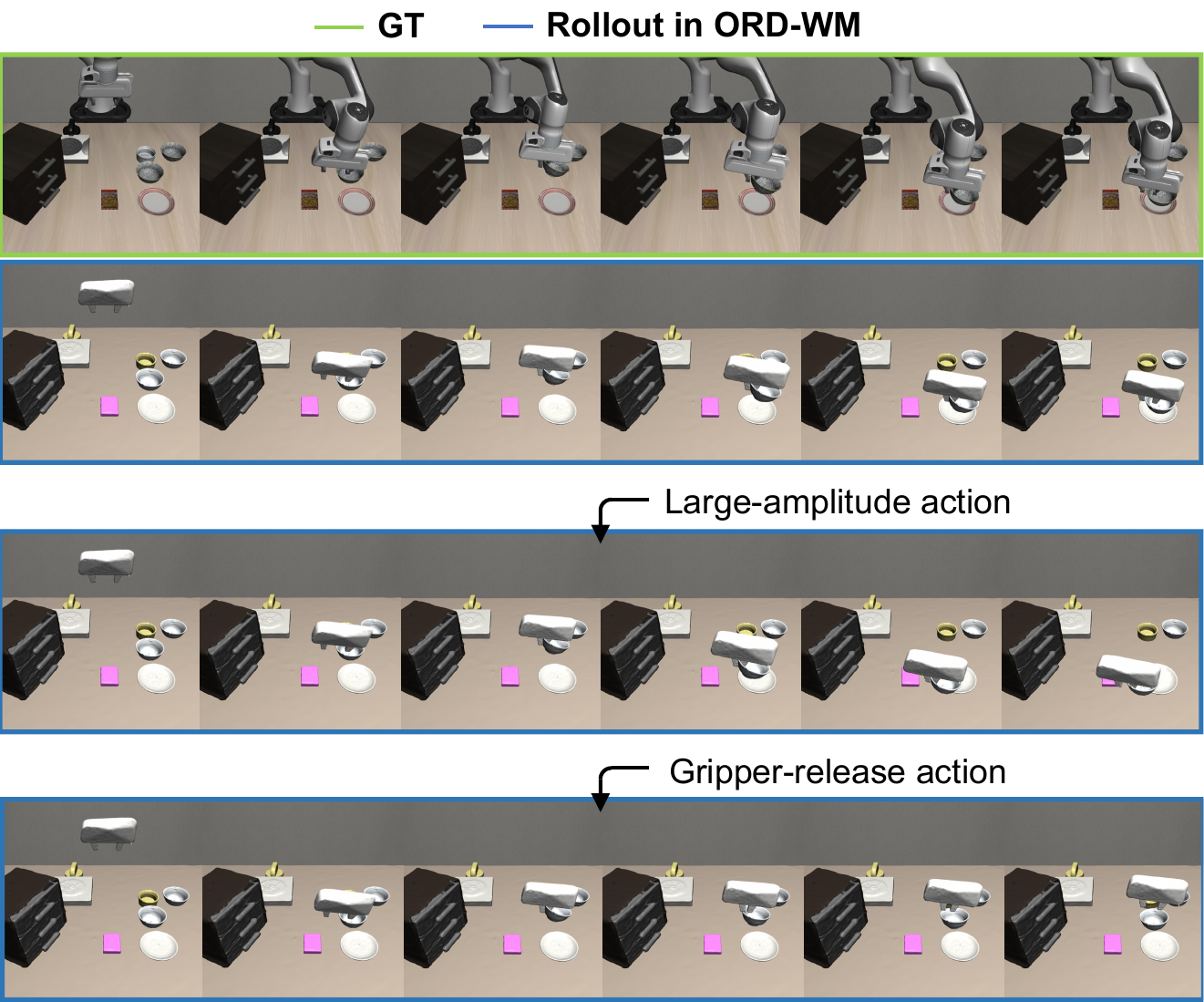}
\caption{Counterfactual action rollouts. Starting from the branching time step indicated by the arrow, ORD-WM is rolled out with a nominal action sequence, a larger-amplitude motion sequence, and a gripper-release sequence. The resulting futures respond differently to the action interventions, showing action-dependent object motion prediction.}
\label{fig:action_conditioned_rollout}
\end{figure}

We further evaluate whether ORD-WM responds to changes in the future action sequence from the same intermediate state. As shown in Fig.~\ref{fig:action_conditioned_rollout}, all rollouts follow the same trajectory until the branching time step indicated by the arrow. Starting from this shared state, we roll out ORD-WM with different future action sequences, including the nominal action sequence, a larger-amplitude motion sequence, and a gripper-release sequence.

The predicted futures change consistently with the action intervention after the branching point. Under the nominal action sequence, the predicted object motion follows the reference behavior. When the motion amplitude is increased, ORD-WM predicts a larger subsequent object displacement. In contrast, when the gripper is released, the object is predicted to lose support from the gripper and drop onto the table. These comparisons show that ORD-WM uses the future action sequence to predict object motion, rather than merely extrapolating a fixed trajectory from the past rollout state.

\subsection{Object Generalization in Rollout}
\label{app:rollout_object_generalization}

We further examine the object generalization behavior of ORD-WM through counterfactual rollouts. Starting from a LIBERO rollout, we replace the original bowl representation with unseen bowl instances from RoboCasa that have different shapes. We then keep the action sequence fixed and roll out ORD-WM from the modified object-centric state. This setting qualitatively evaluates whether the learned dynamics model produces geometry-dependent predictions when the object geometry changes.

\begin{figure}[H]
\centering
\includegraphics[width=0.9\linewidth]{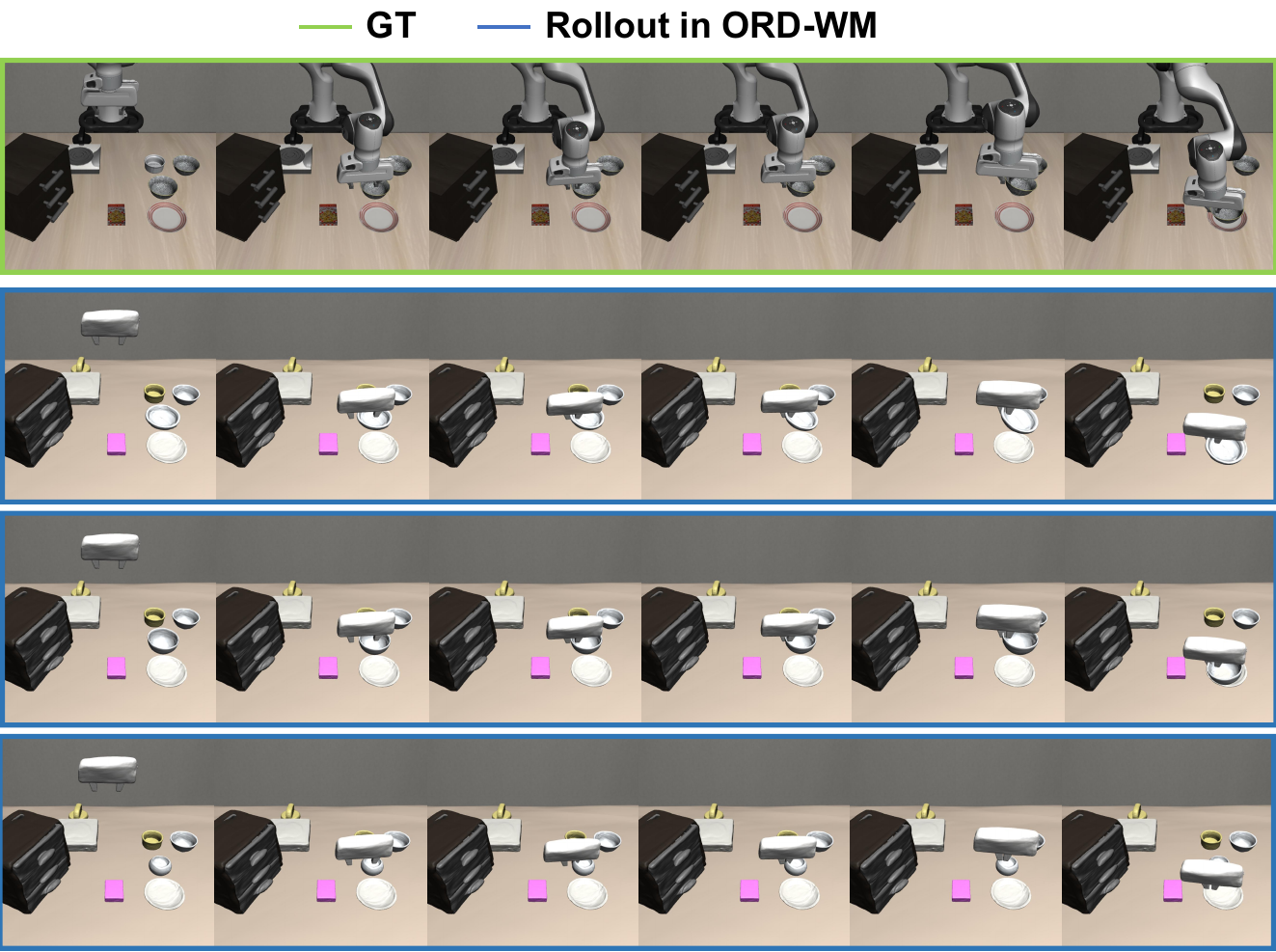}
\caption{Object generalization in counterfactual rollout. We replace the original bowl representation with unseen RoboCasa bowl instances while keeping the same action sequence. ORD-WM produces different rollouts according to the substituted object geometry.}
\label{fig:unseen_obj_rollout}
\end{figure}

As shown in Fig.~\ref{fig:unseen_obj_rollout}, the predicted outcomes vary consistently with the substituted bowl geometry. In the first two examples, ORD-WM predicts that the replaced bowls can be lifted by the gripper under the same action sequence. In the last example, the substituted bowl is not lifted, which is visually consistent with its shape and contact configuration. These counterfactual rollouts suggest that ORD-WM uses the geometry-aware object latent when predicting action-conditioned object motion, rather than simply replaying a fixed motion pattern from the original trajectory.

\subsection{Geometry Preservation During Rollout}
\label{app:geometry_preservation_rollout}

We further analyze the effect of the structured SE(3) update in ORD-WM. To isolate this design choice, we compare ORD-WM with an ablated variant that removes the analytic SE(3) latent update. Instead of predicting an object-wise rigid motion increment and applying it to the current object latent, this variant directly predicts the next latent state for each object. The two models use the same object-centric input representation and rollout setting, but differ in whether object evolution is constrained to follow a rigid transformation in latent space.

\begin{figure}[H]
\centering
\includegraphics[width=\linewidth]{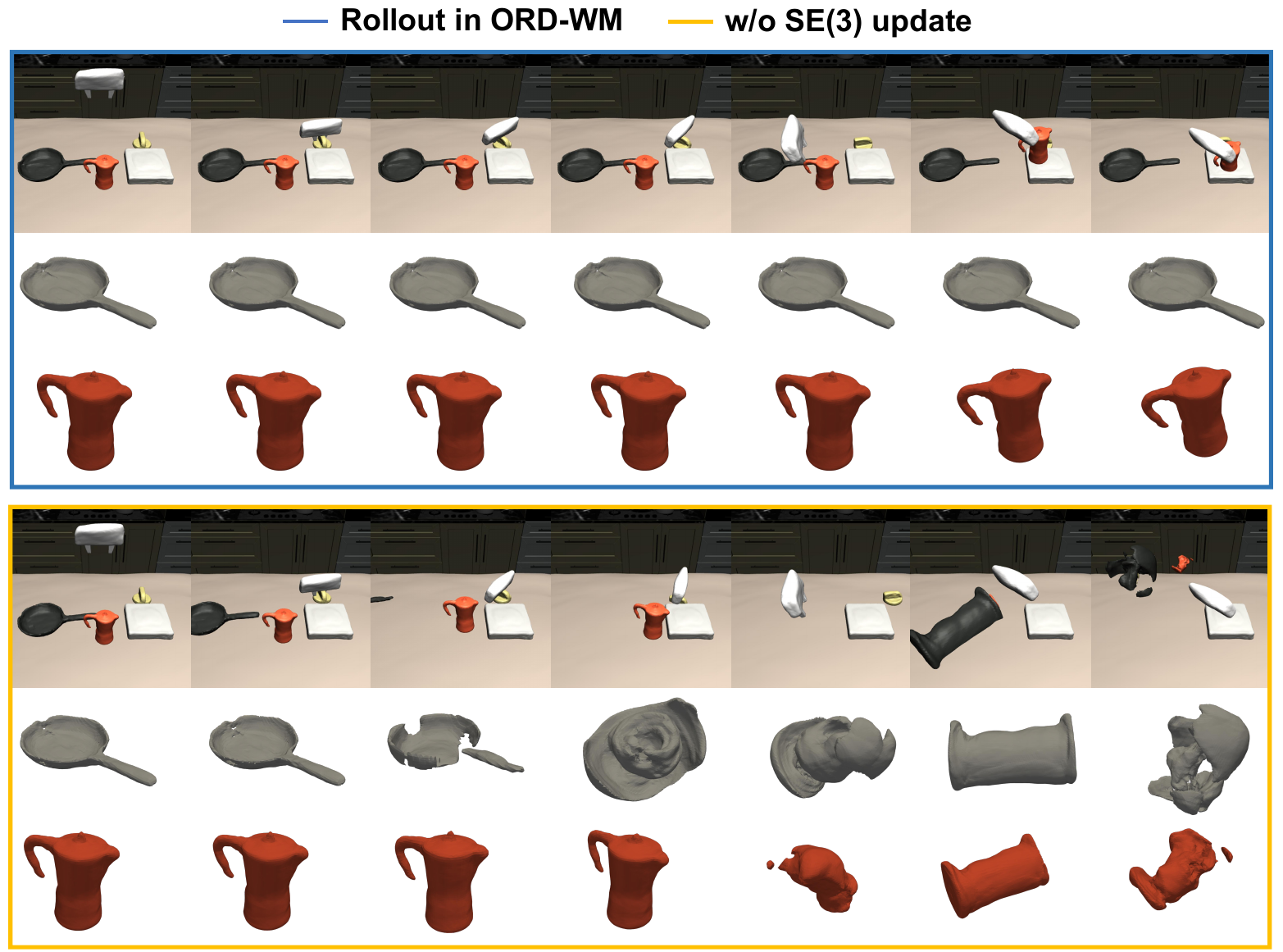}
\caption{Geometry preservation during rollout. ORD-WM preserves object geometry by updating object latents through predicted SE(3) transformations, whereas the variant without the SE(3) update directly predicts future latents and accumulates shape distortion over time.}
\label{fig:rollout_geom}
\end{figure}

As shown in Fig.~\ref{fig:rollout_geom}, ORD-WM preserves object geometry throughout the rollout while updating object poses according to the predicted motion. In contrast, the variant without the SE(3) update gradually introduces geometric distortion when directly predicting future latents. These errors accumulate over time: object shapes first become slightly deformed, then progressively lose their original structure, and eventually collapse into unrealistic geometries. Once the predicted latents deviate substantially from valid object representations, subsequent dynamics prediction also becomes unstable, causing the rollout to diverge. This comparison highlights the importance of applying object-level rigid transformations in latent space rather than freely generating the next object state.

\flushbottom
\section{MPC Implementation Details}
\label{app:mpc}

This appendix provides additional details for the model predictive control (MPC) experiments with ORD-WM. We describe the receding-horizon planning procedure, the CEM optimization settings, and the target-pose cost used for evaluating candidate rollouts.

\subsection{MPC Formulation}
At each control step, the current object-centric scene state, gripper-part states, and predicted object dynamics are used to plan a sequence of future gripper motions. Only floating objects are actively predicted; fixed objects provide static context. The planner uses the Cross-Entropy Method (CEM)~\citep{rubinstein1997optimization} to optimize a sequence of relative gripper motions over a finite horizon \(H\). After optimization, only the first action in the selected sequence is executed, and the robot replans in a receding-horizon manner from updated observations.

The MPC cost compares predicted scene states with target scene states over task-relevant components. For pushing, the cost at rollout step \(h\) is a weighted sum of object position and rotation discrepancies:
\begin{equation}
\mathcal{C}_h = \sum_i \left( w_{\mathrm{pos}} \|\hat c^i_h - c^{i,\star}\|_2 + w_{\mathrm{rot}} d_R(\hat R^i_h, R^{i,\star}) \right),
\end{equation}
where \(c^{i,\star}\) and \(R^{i,\star}\) denote the target position and rotation of object \(i\), and \(d_R(\cdot,\cdot)\) denotes the rotation discrepancy. The total horizon cost includes discounted intermediate steps and a terminal cost:
\begin{equation}
\mathcal{C} = \sum_{h=1}^{H} \gamma^{h-1}\mathcal{C}_h + w_{\mathrm{final}}\mathcal{C}_H.
\end{equation}
Discounting encourages early progress toward the target, while the terminal term emphasizes achieving the final configuration.

\subsection{Implementation Details}
CEM samples action sequences from a diagonal Gaussian distribution in the normalized action space and refits the distribution using the elite samples at each iteration. Sampled actions are clipped to valid bounds before being scaled to physical units.
For pushing, CEM optimizes planar end-effector displacements $a_h=(\Delta x_h,\Delta y_h)$ with fixed orientation and a closed gripper. For pick-and-place, it optimizes three-dimensional displacements $a_h=(\Delta x_h,\Delta y_h,\Delta z_h)$. Candidate displacements are converted into the corresponding gripper-part rigid motions for dynamics prediction. Pick-and-place uses stage-specific target scene states for grasping, lifting, and transport. MPC replans toward the current stage's target. Table~\ref{tab:mpc_settings} summarizes the pushing optimization and cost settings.

\begin{table}[h]
\centering
\caption{Pushing MPC optimization and cost settings.}
\label{tab:mpc_settings}
\begin{tabular}{ll}
\toprule
Item & Value \\
\midrule
\multicolumn{2}{l}{\textit{Optimization settings}} \\
Action space & relative \(xy\) motion \\
Horizon \(H\) & \(8\) \\
Normalized action bounds & \([-1,1]^2\) \\
Physical action range & \(\pm 5\) cm \\
Gripper state & closed \\
Total candidates & \(64 / 256 / 1024\) \\
Candidates / iter. & \(32 / 64 / 256\) \\
CEM iterations & \(2 / 4 / 4\) \\
Elite fraction & \(0.05\) \\
Initial mean & \(0\) \\
Initial std. & \(0.2\) \\
Minimum std. & \(0.03\) \\
Mean momentum & \(0.1\) \\
Std. momentum & \(0.1\) \\
Action clipping & \([-1,1]\) \\
Warm start & no \\
\midrule
\multicolumn{2}{l}{\textit{Cost settings}} \\
Position weight \(w_{\mathrm{pos}}\) & \(5.0\) \\
Rotation weight \(w_{\mathrm{rot}}\) & \(0.5\) \\
Discount \(\gamma\) & \(0.8\) \\
Final weight \(w_{\mathrm{final}}\) & \(5.0\) \\
\bottomrule
\end{tabular}
\end{table}

\end{document}